\documentclass[aoas]{imsart}

\RequirePackage{amsthm,amsmath,amsfonts,amssymb}
\RequirePackage[authoryear]{natbib}
\RequirePackage[colorlinks,citecolor=blue,urlcolor=blue]{hyperref}
\RequirePackage{graphicx}

\usepackage{comment,times}
\usepackage[linesnumbered,ruled]{algorithm2e}
\usepackage{bm}
\usepackage{float}
\usepackage[usenames]{color}
\usepackage[table]{xcolor}
\usepackage{enumitem}
\usepackage{accents}
\usepackage{booktabs}
\usepackage{multirow}
\usepackage{url} 
\usepackage{xr} 
\usepackage{tikz}
\usetikzlibrary{arrows.meta}
\usepackage{orcidlink}

\startlocaldefs
\theoremstyle{plain}

\newtheorem{theorem}{Theorem}[section]

\theoremstyle{remark}

\def\spacingset#1{\renewcommand{\baselinestretch}%
{#1}\small\normalsize} \spacingset{1}

\newcommand{\argmin}{\operatorname*{\arg\min}}

\newcommand{\E}{\operatorname{E}} 

\newcommand {\hF} {\tilde{F}}

\newcommand{\mb}{\boldsymbol}
\newcommand{\mc}{\mathcal}

\usepackage{stmaryrd}

\newcommand{\Z}{\bm Z}
\newcommand{\X}{\bm X}

\newcommand{\z}{\bm z}

\newcommand{\M}{\bm M}

\newcommand{\Er}{\text{R}}
\newcommand{\Gr}{\text{Gr}}

\newcommand{\abs}[1]{\left| #1 \right|}

\endlocaldefs

\begin{document}

\begin{frontmatter}
\title{Boosting Data Analytics with Synthetic Volume Expansion}
\runtitle{Boosting Data Analytics with Synthetic Volume Expansion}

\begin{aug}
\author[A]{\fnms{Xiaotong}~\snm{Shen}\ead[label=e1]{xshen@umn.edu}\orcid{0000-0003-1300-1451}},
\author[A]{\fnms{Yifei}~\snm{Liu}\ead[label=e2]{liu00980@umn.edu}}
\and
\author[B]{\fnms{Rex}~\snm{Shen}\ead[label=e3]{rshen0@stanford.edu}}
\address[A]{School of Statistics, University of Minnesota, Twin Cities\printead[presep={,\ }]{e1,e2}}

\address[B]{Department of Statistics, Stanford University\printead[presep={,\ }]{e3}}
\end{aug}

\begin{abstract}
Synthetic data generation, a cornerstone of Generative Artificial Intelligence, promotes a paradigm shift in data science by addressing data scarcity and privacy while enabling unprecedented performance. As synthetic data becomes more prevalent, concerns emerge regarding the accuracy of statistical methods when applied to synthetic data in contrast to raw data. This article explores the effectiveness of statistical methods on synthetic data and the privacy risks of synthetic data. Regarding effectiveness, we present the Synthetic Data Generation for Analytics framework. This framework applies statistical approaches to high-quality synthetic data produced by generative models like tabular diffusion models, which, initially trained on raw data, benefit from insights from pertinent studies through transfer learning. A key finding within this framework is the generational effect, which reveals that the error rate of statistical methods on synthetic data decreases with the addition of more synthetic data but may eventually rise or stabilize. This phenomenon, stemming from the challenge of accurately mirroring raw data distributions, highlights a ``reflection point''—an ideal volume of synthetic data defined by specific error metrics. Through three case studies, sentiment analysis, predictive modeling of structured data, and inference in tabular data, we validate the superior performance of this framework compared to conventional approaches. On privacy, synthetic data imposes lower risks while supporting the differential privacy standard. These studies underscore synthetic data's untapped potential in redefining data science's landscape. 
\end{abstract}


\end{frontmatter}

\section{Introduction} 
\label{intro}

The advent of synthetic data generation, fueled by generative artificial intelligence (GAI), has initiated a paradigm shift in data analytics, pivoting towards the adoption of synthetic data. Synthetic data, designed to simulate real-world scenarios, presents a viable alternative to the challenges associated with traditional data collection, sharing, and analysis, especially when data are scarce or sensitive. As predicted by Gartner, synthetic data will account for 60\% of the data used in AI and analytics projects by 2024, surpassing the utilization of raw data in AI models by 2030 \citep{gartner, mitreport}. Highlighting its transformative impact, a Forbes article by \cite{toews2022synthetic} predicts that synthetic data technology "will reshape the world of AI in the years ahead, scrambling competitive landscapes and redefining technology stacks."  Furthermore, \cite{jordon2022synthetic} provides an exhaustive review of synthetic data, highlighting its considerable promise as a technology. This evolution signifies a shift away from the traditional reliance on raw data, opening a new chapter in data generation and its application of analytics.


This paper investigates the challenges of efficacy and privacy in synthetic data utilization, emphasizing the role of synthetic data in enhancing data analytics. Specifically, we introduce the Synthetic Data Generation for Analytics (Syn) framework, aimed at increasing the precision of statistical methods applied to high-fidelity synthetic data that accurately replicates raw data through transfer learning. This framework mirrors the statistical properties of raw data through advanced knowledge transfer techniques while offering a potential solution for data sharing without compromising data privacy. Synthetic data provides two main benefits for data analytics: it mitigates data scarcity and addresses privacy issues \citep{ghalebikesabi2023differentiall}.

In the Syn framework, a generative model, informed by raw data, is used to produce synthetic data that replicates its distribution, incorporating insights from pre-trained models in relevant studies through transfer learning. The Syn framework enables the integration of both pre-trained and generative models of various kinds that share comparable latent structures \citep{akbar2023beware}. It employs an array of generative models tailored for different domains: image diffusion \citep{sohl2015deep, wei2021diffusion}, text diffusion models \citep{ho2020denoising}, text-to-image diffusion models \citep{zhang2023text}, time-series diffusion models \citep{lin2023diffusion}, spatio-temporal diffusion models \citep{yuan2023spatio}, and tabular diffusion models \citep{kotelnikov2023tabddpm,zheng2022diffusion, kim2022stasy, 
lee2023codi}. Moreover, Syn embraces advanced models such as the Reversible Generative Models \citep{kingma2018glow}. These flow-based models capture the raw data distribution and can estimate both conditional and marginal distributions. 

  Through Syn exploration, we address a critical issue of efficacy: Can high-fidelity synthetic data boost the effectiveness of statistical methods solely reliant on raw data? If so, how may we implement such enhancements? Recent research offers diverging viewpoints on this issue. In some cases, synthetic X-ray images improve the accuracy of machine learning models \citep{gao2023synthetic}, whereas, in others, training on synthetic data may compromise performance for some machine learning models \citep{kotelnikov2023tabddpm}.

 Our investigation reveals that synthetic data, when accurately generated, can boost the accuracy of a statistical method by augmenting the sample size of raw data. However, a significant caveat emerges: a statistical method applied to low-fidelity synthetic data could yield unreliable outcomes. Often, a generational effect emerges, whereby as the size of synthetic data grows, 
the precision gain of a method may diminish or even plateau. This phenomenon has been illustrated in our study for structured data prediction in Section \ref{prediction}. This challenge arises from generation errors or discrepancies between the data-generation distributions of synthetic and raw data. Fundamentally, the generational effect underscores a key concern: regardless of the size of synthetic data, generation errors can compromise the accuracy of a statistical method.

While evaluating a supervised task is straightforward, hypothesis testing presents the challenge of regulating the Type-I error
while enhancing the power of a test. To address this challenge, we introduce ``Syn-Test,'' a test that augments the sample size of raw data by applying synthetic data. For clarity, ``Syn-A'' denotes Method A within the Syn framework
throughout this article. Syn-Test determines the ideal size of synthetic data required to manage the empirical Type-I error while performing a test for finite inference samples using Monte Carlo methods. Our research indicates that the ideal size of synthetic data can heighten the accuracy. Moreover, our theoretical investigation sheds light on the generational effect, precision, and the size of synthetic data.

Additionally, we introduce Syn-Slm, a streamlined approach enhancing Syn's usability in various applications. This method bypasses the training of conventional predictive models in supervised learning tasks. Instead, it directly models the outcome's conditional distribution through advanced generative models. The effectiveness of Syn-Slm is demonstrated with two examples: sentiment analysis using text data and regression analysis with tabular data.

To highlight the capabilities of the Syn framework, we explore sentiment analysis, predictive modeling for structured data, and tabular data inference. In these domains, methods using high-fidelity synthetic data outperform those with raw data, attributed to the synthetic data generation by diffusion models obtained by fine-tuning relevant pre-trained models on raw data. This fine-tuning boosts efficiency and accuracy by leveraging existing knowledge akin to a scientist solving specialized problems. Specifically, it updates a generative model's architecture, like adjusting a neural network's weights or training the entire network or a subnetwork, with the rest remaining unchanged. Furthermore, one may augment network architectures with adapters to accommodate various input dimensions and tasks. For example, a tabular diffusion model pre-trained on an Adult-Male dataset can be fine-tuned on an Adult-Female dataset with the same features, as discussed in Section \ref{prediction-real}. Similarly, a diffusion model initially trained with X-ray images \citep{akbar2023beware} could undergo fine-tuning, including potential modifications to the network architecture, for CT-scan images. Despite originating from distinct domains, they exhibit visual resemblances.

In the first area, we contrast three models: OpenAI's Generative Pre-trained Transformer (GPT)-3.5 for illustrating Syn-Slm, DistilBERT \citep{sanh2019distilbert}) using transfer learning, and LSTM in the traditional framework for analyzing consumer reviews from the IMDB movie dataset.
In this context, Syn's generative capability using GPT-3.5 significantly outperforms the LSTM
approach in the traditional setup. Yet, DistilBERT in the transfer learning setup, although trailing, demonstrates notable competitiveness compared to GPT-3.5. To further validate Syn-Slm's application, we provide a simulated example with tabular data in Appendix \ref{appendix: syn-slm}.

In the second area, we introduce Syn-Boost, a version of CatBoost \citep{dorogush2018catboost}—a gradient boosting algorithm \citep{schapire1990strength}—trained on synthetic data.
Syn-Boost bolsters the precision of CatBoost for both regression and classification tasks
across eight real-world datasets, utilizing a refined tabular diffusion model \citep{kotelnikov2023tabddpm}. Statistically, the error trajectory of Syn-Boost exhibits either a U-shaped or L-shaped pattern, determined by the generation errors and the volume of synthetic data
used. Moreover,  when employing the same knowledge transfer methods, Syn-Boost surpasses traditional feed-forward networks trained on raw data in six of eight cases. These observations highlight that the generative approach of Syn offers a predictive 
advantage even against a top predictive methodology given the same data input.

In the third domain, we explore feature significance tests in discerning feature relevance in regression and classification using CatBoost. This exploration within black-box models unveils largely uncharted territory. Recently, \cite{dai2024significance} introduces an asymptotic test through sample splitting. To augment its statistical prowess, we employ the Syn-Test, capitalizing on pre-trained generative models and knowledge transfer, as demonstrated in two distinct scenarios. 
In one scenario, we leverage a pre-trained model to ensure smooth knowledge transfer from the male data to the female data while improving generation fidelity and test accuracy for female data, especially when male and female data distributions present distinct characteristics.
These observations emphasize the importance of knowledge transfer in mitigating disparities in domains such as healthcare and social science, particularly when data for specific subgroups, like minorities, are limited. Moreover, we shed light on the ``generational effect'', accentuating the invaluable interplay of synthetic data generation and knowledge transfer in hypothesis testing—a domain 
rarely explored in current literature.

This article consists of six sections. Section \ref{sec: enhacing} explores Syn's role in enhancing the accuracy of statistical methods and emphasizes the importance of knowledge transfer in synthetic data generation.
Section \ref{numerical} provides illustrative examples, exploring the generational effect in predictive modeling and inference. Section \ref{privacy} focuses on the privacy issue of synthetic data.  In Section \ref{future}, we discuss the implications of generating synthetic data for data science. The Appendix contains technical details.

\section{Enhancing Statistical Accuracy}\label{sec: enhacing}

\subsection{Synthetic Data}

\begin{figure}
    \centering
\includegraphics[width=\textwidth]{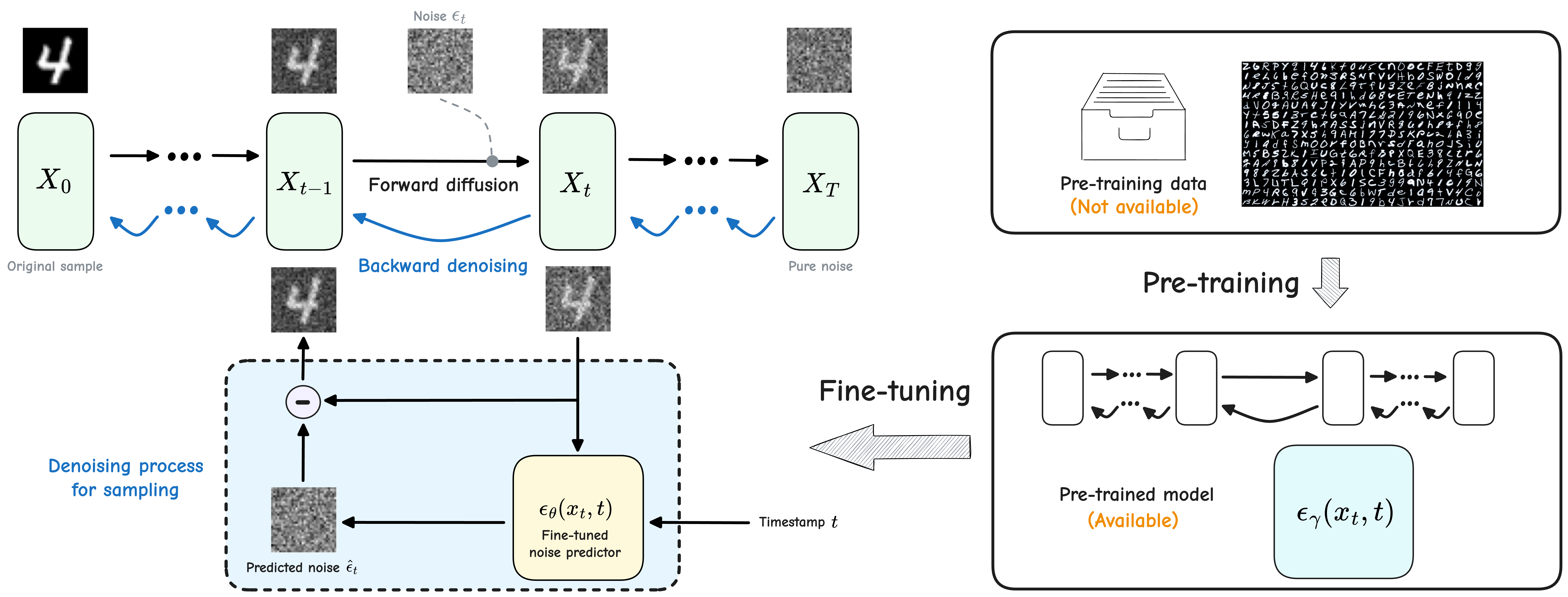}
\caption{Illustration of denoising diffusion probabilistic model \citep{ho2020denoising}.
In the forward process, noise $\epsilon_t$ sequentially corrupts the sample $\X_t$, evolving from the original sample $\X_0$ to a target, such as random noise, over  $t=0,\cdots, T$. Conversely, the backward process employs a neural network $\epsilon_{\theta} (\X_t, t)$ to predict $\epsilon_t$, starting from the random state. This network, fine-tuned from similar pre-trained models,  denoises $\X_t$, from $t = T, ..., 0$, to generate a synthetic sample $\tilde{\X}_0$ replicating $\X_0$.}
    \label{fig: ddpm-illustration}
\end{figure}

The Syn framework empowers data analytics by applying statistical methods on a synthetic sample $\tilde{\Z}^{(m)}=(\tilde{\Z}_i)_{i=1}^m$. This sample is generated by a generative model trained on raw data $\Z^{(n)}=(\Z_i)_{i=1}^n$ through fine-tuning a pre-trained model, leveraging insights from various similar studies.
These models include GPT \citep{brown2020language, bubeck2023sparks, openai2023gpt4}, diffusion models \citep{ho2020denoising, song2020denoising}, normalizing flows \citep{dinh2014nice, dinh2016density, kingma2018glow}, and GANs \citep{frid2018gan, liu2020roundtrip}. 
For an in-depth understanding of the generation processes for diffusion models and flows, 
readers refer to \citep{liu2023perturbation}.
In this framework, the cumulative distribution function (CDF) $\tilde F$ of $\tilde{\Z}^{(m)}$ estimates the CDF $F$ of $\Z^{(n)}$. 
To produce high-fidelity synthetic data, fine-tuning a pre-trained generative model is recommended, which involves transferring knowledge from previous studies. If pre-trained models are unsuitable, constructing a generative model from scratch is a viable, though less preferred, alternative. The quality of the generated data hinges on the choice of generative model and the effectiveness of knowledge transfer from similar studies. For an illustration, we detail the 
synthetic data generation process using a diffusion model in Figure \ref{fig: ddpm-illustration}. Furthermore, to demonstrate the importance of knowledge transfer, we fine-tune a pre-trained tabular diffusion model \citep{shwartz2022tabular} on the Adult-Male dataset to apply this knowledge to the Adult-Female dataset in Section \ref{feature}, where male and female distributions exhibit distinct differences. For a detailed explanation of the impact of knowledge transfer, 
readers refer to Section \ref{transfter}.

It is imperative to underscore the pivotal role of knowledge encapsulated in pre-trained models for improving generation accuracy through fine-tuning. 
Nevertheless, directly accessing the pre-training data for pre-trained models is often infeasible, impeded by privacy concerns, extensive storage needs, and data inconsistencies \citep{sweeney2002kanonymity, quionero2009dataset, labrinidis2012challenges, carroll2006movement}, 
as seen in models like GPT-4. Furthermore, the distributions of pre-training datasets, especially from different sources, might not always mirror that of raw data,
as illustrated by the Adult-Male and Adult-Female examples in Section \ref{feature}. 
Considering these challenges and the nuances of real-world scenarios, we omit pre-training data from our raw data composition throughout this article.

To yield $\tilde F$ directly, one can utilize some revertible generative models such as normalizing flows and Roundtrip GAN \citep{liu2020roundtrip}, acting as a nonparametric estimate of $F$.
For other generative models, such as diffusion models and GPT, one can typically obtain $\tilde F$ from synthetic data employing Monte Carlo methods, which we elaborate on in Section \ref{dist}.
In the numerical examples presented in this paper, we utilize a tabular diffusion model (TDM, \cite{kotelnikov2023tabddpm}) and 
GPT for synthetic data generation.


Subsequently, we explore the precision advantages offered by the Syn framework.

\subsection{Optimal Synthetic Size for Estimation and Prediction}\label{sec:prediction-method}

In estimation and prediction, leveraging a statistical method on a synthetic sample gives rise to an estimate, denoted as $\widehat{\bm \theta}(\tilde{\Z}^{(m)})$, of a parameter vector $\bm \theta$. The effectiveness of this method gauges through a specific error metric $\Er(\widehat{\bm \theta}(\tilde{\Z}^{(m)}))=\E L(\widehat{\bm \theta}(\tilde \Z^{(m)}),
\bm \theta)$, which would theoretically improve the statistical accuracy of $\widehat{\bm \theta}(\tilde{\Z}^{(m)})$ with an infinite amount of synthetic data, provided $\tilde F$ perfectly replicates $F$. Here, $\E$ symbolizes the expectation under $F$, while $L(\cdot,\cdot)$ represents a loss function quantifying the discrepancies between $\widehat{\bm \theta}$ and $\bm \theta$. However,
numerical insights from Section \ref{prediction} reveal the existence of a reflection point, denoted as $m_0=\arg\min_{m \geq 1} \Er(\widehat{\bm \theta}(\tilde \Z^{(m)}))$, which delineates a relationship between the synthetic sample size $m$ and the accuracy augmentation for this method. This point $m_0$ is governed by the generation error measured by metrics such as the total variation distance between $\tilde F$ and $F$, defined as $\text{TV}(\tilde F, F)=
\sup_{B}|P_F(B)-P_{\tilde F}(B)|$, where $P_F$ and $P_{\tilde F}$
are probabilities measures induced by $F$ and $\tilde F$ and $B$ is any event.

 To estimate $m_0$, we optimize its empirical risk measure across $m$ on an independent cross-validation sample from the original resources, which yields an optimizer $\hat m$ as an estimate of $m_0$. For instance, in scenarios where the risk measure is the generalization error in binary classification, its empirical risk is the test error on a test dataset obtained from the original resources, approximating a classifier's generalizability.

To investigate the theoretical aspects of the generational effect concerning the accuracy of $\widehat{\bm \theta}(\tilde \Z^{(m)})$, we clarify the notation: Let $\Gr^{(m)}=\Er(\widehat{\bm \theta}(\tilde \Z^{(m)})) - \Er(\widehat{\bm \theta}(\Z^{(m)}))$ represent the discrepancy between the synthetic and raw errors for a sample size $m$, where $\Er(\widehat{\bm \theta}(\Z^{(m)}))=\E L(\widehat{\bm \theta}(\Z^{(m)}))$ denotes the risk incurred when employing a raw sample $\Z^{(m)}$ of size $m$.

\begin{theorem}[Reflection Point]
\label{thm1}
Suppose $\Er(\widehat{\bm \theta}(\Z^{(m)})) = C_{\mb \theta} m^{-\alpha}$ for some constant $\alpha > 0$.
Assume that $\Gr^{(m)} \geq m f(\text{TV}(\tilde F, F))$ if $m \leq m^*$ and $\Gr^{(m)} \geq \Gr^{(m^*)}$
if $m > m^*$ for some finite index $m^*$, where $f(\cdot)>0$ is a function and $f(\text{TV}(\tilde F, F))$ indicates the magnitude of generation error. Then,
\begin{enumerate}
    \item[(i) ] The optimal synthetic size $m_0 < m^*$ is finite.
    \item[(ii)] For any $m > m^*$, $\Er(\widehat{\mb \theta}(\tilde \Z^{(m)})) > \Er(\widehat{\mb \theta}(\tilde \Z^{(m_0)}))$ 
provided that $f(\text{TV}(\tilde F, F))$ is larger than  $\Er(\widehat{\mb \theta}(\tilde \Z^{(m_0)}))/ m^*$.
\end{enumerate}
\end{theorem}


The premise regarding the growth of $Gr^{(m)}$ in $m$ in Theorem \ref{thm1} suggests that the divergence between the synthetic risk and the risk escalates more rapidly when $m \leq m^*$, beyond which it remains higher than $Gr^{(m^*)}$. This condition implies a substantial generation error. According to Theorem 1, a considerable generation error results in the synthetic risk $\Er(\widehat{\mb \theta}(\tilde \Z^{(m)}))$  reaching its minimum at a specific $m_0$, beyond which the synthetic error begins to worsen. This outcome indicates that an increase in synthetic sample size does not necessarily improve estimation or prediction accuracy in the presence of significant generation error, a phenomenon further illustrated in Section \ref{prediction}. In contrast, when the generation error is minimal, the synthetic risk remains controlled, and the optimal $m_0$ tends to be substantially large or potentially infinite, as suggested by Theorem \ref{thm2}.

Next, we establish a bound on the synthetic risk to offer insight into the conditions under which accuracy improvements may arise.
Assume that $\Z^{(n)}$ is independently and identically distributed according to $F$ while $\tilde \Z^{(m)}$ is independently and identically distributed according to
a conditional distribution $\tilde F \equiv F_{\Z \mid \Z^{(n)}}$ given $\bm Z^{(n)}$.

\begin{theorem}[Accuracy Gain] \label{thm2}
Let $L$ be a nonnegative loss function upper-bounded by $U>0$. For any $m \geq 1$,  
\begin{equation}\label{cor_eq_1}
\Er(\widehat{\bm \theta}(\tilde \Z^{(m)})) \leq \Er(\widehat{\bm \theta}(\Z^{(m)})) + 2 U m \text{TV}(\tilde F,F). 
\end{equation}
Moreover, if $\Er(\widehat{\bm \theta}(\Z^{(m)})) = C_{\mb \theta} m^{-\alpha}$, then
\begin{equation}\label{cor_eq_2}
\Er(\widehat{\bm \theta}(\tilde \Z^{(m_0)})) \leq \Er(\widehat{\bm \theta}(\tilde \Z^{(s_0)})) \leq
\left(\alpha^{\frac{-\alpha}{1 + \alpha}} + \alpha^{\frac{1}{1 + \alpha}} 
\right) (2U)^{\frac{\alpha}{1 + \alpha}} C_{\mb \theta}^{\frac{1}{1 + \alpha}} \cdot 
\text{TV}( \tilde F,F) ^{\frac{\alpha}{1 + \alpha}},
\end{equation}
when $s_0 = \left (\frac{2U}{\alpha C_{\mb \theta}}\right )^{\frac{\alpha}{1 + \alpha}} \cdot \text{TV}(\tilde F,F)^{-\frac{1}{1 + \alpha}}$. Hence,
$\Er(\widehat{\bm \theta}(\tilde \Z^{(m_0)})) \leq \Er(\widehat{\bm \theta}(\Z^{(n)}))$ achieves an accuracy gain under Syn when 
the total variation $\text{TV}(\tilde F,F)$ is sufficiently small. For example, this occurs when $\text{TV}(\tilde F,F) \leq C_{\mb \theta} (2U)^{-1} \left( \alpha^{\frac{-\alpha}{1+\alpha}} + \alpha^{\frac{1}{1 + \alpha}} \right)^{-\frac{1 + \alpha}{\alpha}} \cdot n ^{-(1 + \alpha)}$.
\end{theorem}

Theorem \ref{thm2} posits that training a method trained on synthetic data can lead to an accuracy gain, provided that the generation error
that governs the synthetic risk $\Er(\widehat{\bm \theta}(\tilde \Z^{(m)}))$ is small. Hence, high-fidelity data can mitigate the synthetic risk in a method, thereby enabling a large optimal synthetic size $m_0$ for further amplifying the precision.

The existing literature provides insights into the magnitude of generation error as represented by $\text{TV}(\tilde F, F)$. For instance, Theorem 5.1 in \citet{oko2023diffusion} specifies bounds for a diffusion model, given the data-generating distribution is a member of a Besov space. It's worth noting that the boundary defined in \citet{oko2023diffusion} pertains to $\E \text{TV}(\tilde F, F)$, with $\E$ symbolizing the expectation relative to $F$. However, one may extend this to the in-probability convergence rate for the random quantity $\text{TV}(\tilde F, F)$ by leveraging Markov's inequality, which provides the convergence rates for $\text{TV}(\tilde F, F)$ based on the raw sample size $n$ and/or the pre-training sample size.

\subsection{Optimal Synthetic Size for Hypothesis Testing} \label{sec: hypothesis-testing}

 We now introduce Syn-Test, a novel inference tool using high-fidelity synthetic data to boost any test's power by expanding 
the sample size of raw data. Syn-Test yields two distinct advantages. First, it employs synthetic data to gauge the null distribution of any test statistic by Monte Carlo methods as in the bootstrap approach \citep{efron1992bootstrap}, circumventing analytical derivations. This methodology proves particularly powerful for unstructured data inferences, including texts and images \citep{liu2023perturbation}. Second, Syn-Test identifies the optimal synthetic data size, optimizing a test's power while maintaining a suitable control of Type-I errors. For illustration, we refer to Section \ref{feature}.

Given a raw sample, Syn-Test employs two nearly equal-sized subsamples $\mathcal S_1$ and $\mathcal S_2$, partitioned from a training sample for fine-tuning a pre-trained generative model. It also uses a separate validation sample $\mc S_3$ for validating model training. One generative model generates synthetic data using $\mathcal S_1$ for null distribution estimation, while the other uses $\mathcal S_2$ for computing the test statistic. Figure \ref{fig:syntest} illustrates the splitting scheme and Syn-Test process.
Syn-Test also empirically determines the optimal synthetic data size $m_0$ to control the Type-I error. By swapping the roles of $\mathcal S_1$ and $\mathcal S_2$, Syn-Test can de-randomize the partition, transitioning the original inference sample size $n$ to a synthetic inference sample size of $m$. This swapping mechanism proves especially advantageous in scenarios with low generation error. Crucially, abundant synthetic data can enhance the size of the raw data,  even when sample splitting results in a reduced inference sample size \citep{wasserman2009high, wasserman2020universal}.

Syn-Test encompasses four steps, using a significance level 
$\alpha$, a tolerance error $\varepsilon$,  and a Monte Carlo size 
$D$.
Syn-Test goes as follows:

\textbf{Step 1: Controlling Type-I Error.} Generate $D$ distinct synthetic samples $(T(\tilde{\Z}_1^{(m,d)}))_{d=1}^D$ of size $m$ by refining a pre-trained generative model with $\mathcal S_1$ under $H_0$, using $\mc S_3$ as a 
validation set for model tuning to avoid overfitting. Compute the empirical distribution of the test statistic $T$ using $(T(\tilde{\Z}_1^{(m,d)}))_{d=1}^D$. Define a rejection region $C_m$ at a significance level where $\alpha>0$.

\textbf{Step 2: Optimizing Synthetic Size through Tuning.}
Execute Step 1, but use $\mathcal S_2$ instead of $\mathcal S_1$ to produce $\tilde{\Z}^{(m,d)}_2$. Utilize the empirical distribution from $(T(\tilde{\Z}_2^{(m,d)}))_{d=1}^D$ to find the empirical Type-I 
error, denoted $\tilde P(C_m)$, for the $C_m$ created in Step 1. 
To effectively control the Type-I error, we propose two distinct strategies for identifying $\hat m$:  an aggressive and a conservative approach. The aggressive approach selects the largest $m$ that maintains the estimated Type-I error within the desired limit. In contrast, the conservative one chooses the smallest $m$ about failing to control the estimated Type-I error. Mathematically,
 
\begin{enumerate}
 \item Aggressive: $\hat m =\max \{m: \tilde P(C_m) \leq \alpha + \varepsilon\}$.
\item Conservative: $\hat m = \min \{m: \tilde P(C_m) \leq \alpha + \varepsilon\ \text{ and } \tilde P(C_{m + 1}) > \alpha + \varepsilon\}$.
\end{enumerate}

In practice, we recommend adopting a conservative approach, as it more effectively manages Type-I errors, although it may be 
less powerful compared to a more aggressive strategy.

\textbf{Step 3: Calculating the P-value.} With the determined $\hat m$, produce synthetic data $\tilde{\Z}_2^{(\hat m)}$ by fine-tuning the pre-trained generative model using $\mathcal S_2$ with $\mc S_3$ as 
a validation set. Calculate the test statistic for $T(\tilde \Z_2^{(\hat m)})$ and determine the P-value, $P^1$, leveraging the null CDF estimated from $\mathcal S_1$ in Step 1.

\textbf{Step 4: Combining the P-values.} 
Repeat Step 3 by interchanging the roles of $\mathcal S_1$ and $\mathcal S_2$ 
to compute the P-value $P^2$. Combine P-values via Hommel's weighted average \citep{hommel1983tests}:
$$\bar{P} = \min \Big(C \min_{1 \leq q \leq 2}\frac{2}{q}
P^{(q)},1\Big),
$$
where $C=\sum_{q=1}^2 \frac{1}{q}=3/2$ and $P^{(q)}$ is the $q$-th order statistic of $\{P^{1},P^{2}\}$.

Hommel's method excels in controlling the Type-I error relative to many of its peers, ensuring that $\mathbb{P} \big(\bar{P} \leq \alpha\big) \leq \alpha$ under $H_0$. While effective, there are also alternative strategies such as the Cauchy combination \citep{liu2020cauchy}. To expedite the search of an estimated $m_0$ $\hat m$, we may consider techniques such as Bisection \citep{burden2001numerical} or Fibonacci Search \citep{kiefer1953sequential}.

\begin{theorem}[Validity and power of Syn-Test]
\label{thm3} 
Assume that $\tilde \Z^{(m, d)}$ is an i.i.d. sample of size $m$ following $\tilde F=F_{\tilde \Z \mid \Z^{(n)}}$ given $\Z^{(n)}$. Let $\hF_T$ and $F_T$ be the synthetic and raw distributions of $T$ calculated on a sample of size $m$. Then,  the estimation error of the null distribution is governed by the Monte Carlo error and the generation error:      
\begin{equation} 
\label{bounds}
 \sup_{x} \abs{F_{\tilde T}(x) - F_T(x)} \leq \sqrt{\frac{\log \frac{2}{\delta}}{2D}} +\text{TV}(\tilde{\Z}^{(m)}, \Z^{(m)}).
\end{equation}                
As a result, Syn-Test offers a valid test as long as $\text{TV}(\tilde{\Z}^{(m)}, \Z^{(m)}) = m \cdot \text{TV}(\tilde F, F)\rightarrow 0$
\footnote{Assuming $\tilde F$ is derived from pre-training data of size $N$, then $m \text{TV}(\tilde F, F)$ decreases as $N \rightarrow \infty$, given certain regularity conditions. To meet the requirement, $\text{TV}(\tilde F, F)$ must reduce at a rate quicker than the growth of $m$, underscoring the significance of extensive pre-training data size, a common practice in transfer learning.} in probability and $D \rightarrow \infty$.
Moreover, let the power function $\phi_{m, \alpha}$ be $P(T(\Z^{(m)}) \in R_\alpha | H_a)$ for rejection region $R_\alpha$ at significance level $\alpha$, and $\tilde \phi_{m, \alpha}$ analogously with $\tilde \Z^{(m)}$. If for some $m > n$, $\Delta = \phi_{m, \alpha} - \phi_{n, \alpha} > 0$, then $\tilde \phi_{m, \alpha} > \phi_{n, \alpha}$ when $\text{TV}(\tilde F, F) < \Delta / m$, indicating that Syn-Test enhances power if the generation error is small.
\end{theorem}

Syn-Test facilitates valid inference without necessitating many model assumptions, specific data distributions, or an infinitely large inference sample. Its validity and power primarily hinge on the small generation error, a condition usually met by generative models trained on sufficiently large datasets. Additionally, Syn-Test permits a fine-tuned generator to estimate the bias of a test statistic through a Monte Carlo approach. This estimation can serve various purposes in downstream analysis, including debiasing. We defer this methodology in the future study.

\subsection{Syn-Slm: Streamlined Approach} \label{dist}

The Syn Framework enables synthetic data generation, mirroring raw data distributions. It is adept at tackling statistical challenges in both unsupervised and supervised realms. It can also derive $\tilde F$ directly from some generative models such
as normalizing flows \citep{dinh2014nice, dinh2016density, kingma2018glow}.
Subsequently, we introduce Syn-Slm, a streamlined approach for supervised learning tasks. Unlike the methods discussed in Section \ref{sec:prediction-method}, Syn-Slm directly models the conditional distribution of the outcome given
predictors without relying on an additional predictive model for synthetic data. This method simplifies the process by directly modeling the conditional distribution of the outcome.

Supervised learning aims to predict an outcome based on a set of predictors, denoted as $\X$. Consider a scenario with a one-dimensional outcome variable $Y$ and define $\Z=(Y,\X)$. In this context, the goal is to estimate a statistical functional, represented as $\phi(F_{Y | \X})$, where $F_{Y | \X}$ is the conditional CDF of $Y$ given $\X$. Our Syn-Slm method introduces a plug-in estimate $\phi(\tilde F_{Y | \X})$, where $\tilde F_{Y | \X}$ represents the conditional CDF of a synthetic $Y$ given $\X$. This approach, focusing on estimating $\tilde F_{Y | \X}$, is notably distinct from conventional methods, which often estimate $\tilde F_{\X | Y}$, such as generating images from a specific class. However, as hypothesized by \cite{zhang2023mixed}, this type of generative modeling could outperform direct predictive modeling, as demonstrated in their work using imputation techniques. To illustrate, if $\phi(F_{Y | \X}) = \E (Y | \X)$ signifies the conditional expectation of $Y$, then $\phi(\tilde F_{Y | \X}) = \E (\tilde Y | \X)$, where $\tilde Y$ is derived from $\tilde F_{Y | \X}$, resulting in a Syn-Slm estimate. An example is provided in Section \ref{text}, and a simulated example is presented in Appendix \ref{appendix: syn-slm}.

In contrast, the Syn prediction method described in Section \ref{sec:prediction-method} employs a two-step process: it initially generates joint synthetic data $\tilde \Z = (\tilde Y, \tilde \X)$, followed by training a predictive model using $\tilde \Z$. For instance, Syn-Boost in \ref{prediction} utilizes TDM \citep{kotelnikov2023tabddpm} as the generator and CatBoost \citep{dorogush2018catboost} for prediction. While straightforward in implementation, methods akin to Syn-Boost may incur prediction errors from predictive modeling and generation errors from generating $\Z$. In contrast, Syn-Slm directly models the conditional generation of $Y | \X$ and estimates $\tilde F_{Y | \X}$ via a Monte Carlo approach, offering a fully non-parametric solution. Additionally, Syn-Slm enables the estimation of various statistical functionals $\phi(F_{Y | \X})$, such as $E(Y | \X)$, $\text{Var}(Y | \X)$, and quantiles of $Y | \X$. In contrast, methods like Syn-Boost require training different predictive models under varying assumptions to estimate these quantities. A detailed comparison between Syn-Boost and Syn-Slm is presented in Table \ref{tab:compare-boost-slm}.

\begin{table}
\centering
\caption{Comparison between Syn-Boost and Syn-Slm in supervised learning.}
\begin{tabular}{@{}ccc@{}}
\toprule
                                 & Syn-Boost                                                                                            & Syn-Slm                                                                                                           \\ \midrule
Generative modeling              & Joint distribution of $(Y, \X)$                                                                      & Conditional distribution of $Y | \X$                                                                              \\ \midrule
Estimation of $\phi(F_{Y | \X})$ & Predictive modeling on $(\tilde Y, \tilde X)$                                                          & MC computation with $\tilde Y | \X$                                                                               \\ \midrule
Pros                             & \begin{tabular}[c]{@{}c@{}}Generative modeling is easy;\\ Prediction is fast.\end{tabular}           & \begin{tabular}[c]{@{}c@{}}Fully non-parametric;\\ Estimate different $\phi$ once and for all.\end{tabular}       \\ \midrule
Cons                             & \begin{tabular}[c]{@{}c@{}}Additional prediction error;\\ Needs to tune synthetic size.\end{tabular} & \begin{tabular}[c]{@{}c@{}}Generative modeling is challenging;\\ Prediction relies on MC simulation.\end{tabular} \\ \bottomrule
\end{tabular}%
\label{tab:compare-boost-slm}
\end{table}

\subsection{Generative Model and Knowledge Transfer} \label{transfter}


Knowledge transfer elevates generation accuracy by infusing task-specific generative models with pre-trained knowledge from relevant studies. From the perspective of dimension reduction, we dissect knowledge transfer in two scenarios. In the first situation, consider a generative model $g_{\mb \theta}$ parametrized by $\mb \theta$. Originally trained on an extensive dataset for a generation task, the model undergoes subsequent fine-tuning on a smaller but similar dataset to account for distribution shift, resulting in model $g_{\mb \theta^\prime}$, where the architecture remains consistent across both models, with the essence of knowledge transmission occurring via the transition from $\mb \theta$ to $\mb \theta^\prime$ amid the fine-tuning. In the other situation, a robust pre-trained model undergoes training across multiple tasks, characterized as $(f_1, \ldots, f_t) \circ h$. Here, $f_i$ defines the output function tied to the $i$-th task, $h$ is the shared representation function, and $\circ$ denotes functional composition. Given a learned representation $h$, one only fine-tune $f_0$ during its optimization phase for $f_0$ \citep{tripuraneni2020theory}. As the generative model refines its precursor, it absorbs the precisely calibrated representation $h$. This knowledge transfer thus can augment the generation precision through fine-tuning with a heightened accuracy of the learned $f_0$, facilitating its dimension reduction. It is pivotal to acknowledge that within this configuration, $f_0 \circ h$ and $f_i \circ h$; $i = 1, \ldots, t$, only share the same architecture in $h$. An alternate strategy entails concurrent fine-tuning of  $f_0$ and $h$ to derive a representation explicitly for  $f_0$.

The Syn framework capitalizes on knowledge transfer to bolster its overall efficacy, streamlining the synthetic data generation process. For a visual representation of how a pre-trained generative model is fine-tuned on a specific dataset to facilitate knowledge transfer, refer to Figure \ref{fig: ddpm-illustration}. In Section \ref{feature}, we illustrate knowledge transfer in generative models using a pre-trained model based on adult male data \citep{adult}, subsequently fine-tuned with adult female data for downstream analysis. As demonstrated in Figures     
\ref{fig: dist_diff_gender}, \ref{fig: corr_audlt_gender}, and
Table \ref{tab:distribution-distance},
the fine-tuned model adeptly captures the data distribution, even with a limited size of raw samples.




\section{Case Studies}
\label{numerical}

\subsection{Sentiment Analysis} \label{text}

This subsection presents sentiment classification applied to the benchmark dataset, IMDB \citep{maas2011learning}. This task involves assigning emotions expressed in the text into positive or negative sentiments based on the opinions reflected in each text. The dataset comprises $50,000$ polarized movie reviews,  categorized as ``positive'' or ``negative'' sentiments. These labels correspond to movie scores below four or above seven out of ten, where no movie has more than 30 reviews to prevent significant class imbalance. Here, we use $49,000$ of these reviews as our training data, reserving $1,000$ reviews for testing. We compare Syn's generative approach against its conventional counterpart in a downstream 
prediction task, utilizing three models, GPT-3.5, DistilBERT, and LSTM models \citep{chen2023robust}. 

GPT-3.5 functions primarily as a text completion model, predicting the succeeding token as the sentiment label. Although essentially a completion model, GPT-3.5 is a conditional generative model that aligns with Syn-Slm, which we adapt for predictive tasks.
In contrast, DistilBERT generates a fixed-size embedding of a review, which is then passed to an appended classification head to deduce sentiment likelihood. Unlike GPT-3.5's token generation approach, DistilBERT's technique aligns more closely with traditional predictive modeling with transfer learning for supervised tasks.
Additionally, we train a traditional LSTM model from scratch, eschewing prior knowledge. Like DistilBERT, the LSTM processes an embedding and feeds it into a classification head, rendering it a predictive model.

Table \ref{tab: GPT} compares GPT-3.5, DistilBERT, and LSTM in six performance metrics. The extensive collection of pre-trained models likely contributes to GPT-3.5's outstanding performance. On the other hand, LSTM's underperformance stems from its inability to transfer knowledge. Knowledge transfer plays a crucial role in model performance. For details on training configurations, refer to the Supplementary Materials.

\begin{table}
\centering
\caption{Performance comparison of fine-tuned GPT-3.5 (Syn-classification with knowledge transfer), DistilBERT (Classification with knowledge transfer), and traditional LSTM (Classification without knowledge transfer) on IMDB sentiment analysis. Metrics evaluated include Accuracy, Precision, Recall, Area Under ROC, Area Under PRC, and F1-Score, all assessed on an independent test set.}
\label{tab: GPT}
\resizebox{\columnwidth}{!}{%
\begin{tabular}{lccccccc}
\hline
Model   & Training approach              & Accuracy & Precision & Recall & AUROC & AUPRC & F1-score \\ \hline
GPT-3.5     & Fine-tuning, generative model     & 0.970    & 0.967     & 0.975  & 0.991 & 0.989 & 0.971    \\
DistilBERT & Fine-tuning, predictive model    & 0.939    & 0.930     & 0.954  & 0.985 & 0.984 & 0.942    \\
LSTM       & Training, predictive model & 0.854    & 0.885     & 0.819  & 0.939 & 0.943 & 0.851    \\ \hline
\end{tabular}%
}
\end{table}

\subsection{Prediction for Structured Data} \label{prediction}
This subsection investigates the generational phenomenon and challenges associated with enhancing
the precision of gradient-boosting for regression and classification \citep{breiman1997arcing, friedman2002stochastic}.  It also focuses on the implications for the quality of synthetic data generation.
Within the Syn framework, we designate the boosting method tailored for synthetic data as Syn-Boost. Despite the surge of diverse predictive models, the capabilities of the Syn framework in predictive modeling remain largely untapped. To highlight this potential, we draw contrasts between Syn-Boost and its traditional supervised counterparts: specifically, the boosting algorithm — CatBoost \citep{dorogush2018catboost} and FNN — a fully connected neural network that leverages insights from a pre-trained model, both of which are 
traditional. 
Syn-Boost presents a strategy to harness knowledge transfer in boosting, effectively addressing the transfer learning challenge inherent to the boosting method.

\subsubsection{Real-Benchmark Examples}\label{prediction-real}

To closely emulate real-world scenarios, we investigate situations where available pre-trained models have incorporated insights from relevant studies. For this study, we utilize five classification and three regression benchmark datasets \citep{kotelnikov2023tabddpm}, each encompassing three subsets: pre-training, raw, and test data.
A detailed description of these datasets can be found in Table \ref{tab:dataset-meta-data}.

In the Syn-Boost framework, we utilize a tabular diffusion model, TDM \citep{kotelnikov2023tabddpm}, to generate synthetic data of mixed types that closely match the distribution of the original data. TDM employs multinomial and Gaussian diffusion processes to simulate categorical and continuous attributes. The procedure starts by training a TDM model on pre-existing data and then fine-tuning it with raw data. Subsequently, we use CatBoost on the synthetic data of size $m$, created by TDM for classification and regression tasks. To identify the best $m$ for Syn-Boost's synthetic data, we evaluate the error relative to the synthetic-to-raw data ratio, 
ranging from $1$ to $30$, with a step size of $1$.

For FNN, we engage in transfer learning, starting with pre-existing data and later fine-tuning with raw data. This technique is consistent with TDM's training approach, ensuring that both models have a harmonized foundation for effective knowledge transfer.

\vspace{3pt}
\noindent\textbf{Knowledge Transfer from Identical Distributions.} 
For each dataset mentioned, we utilize pre-training data to train pre-trained models, enabling the downstream knowledge transfer on raw data with either Syn-Boost or FNN.
We use the test data solely for evaluating performance. It is important to note that in this study, the pre-training and raw data used for fine-tuning share the same distribution.

\begin{figure}
\makebox[\textwidth]{
\hspace{-0.9pt}
\includegraphics[scale=0.15]{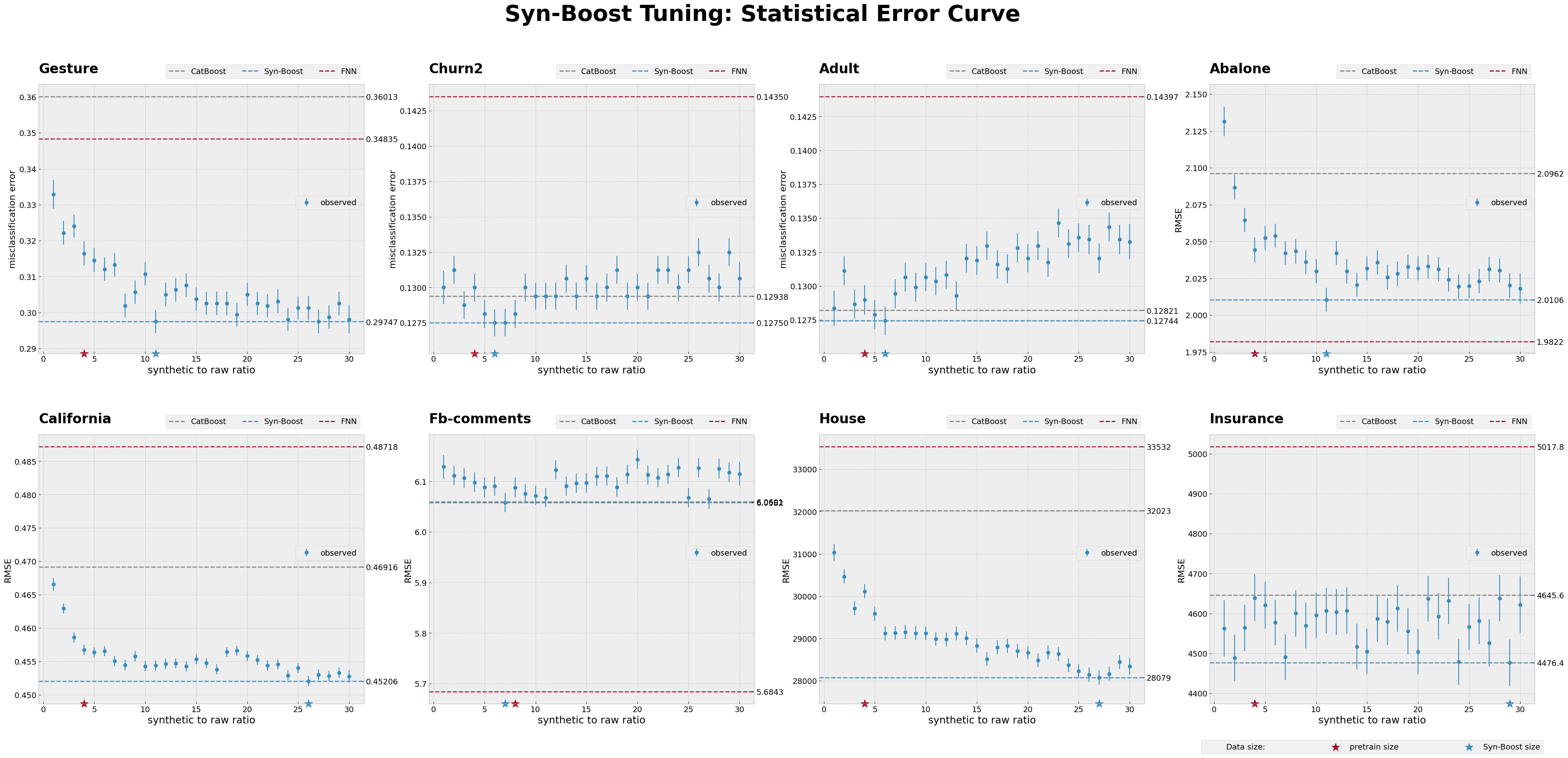}
}
\caption{Comparative error analysis of CatBoost, Syn-Boost, and FNN, with Syn-Boost and FNN applying transfer learning with the same distributions across eight benchmarks \citep{kotelnikov2023tabddpm}, measured at various synthetic-to-raw data ratios. The stars indicate the size of the pre-training data used to obtain pre-trained models and the tuned sample size for Syn-Boost. The performance for classification and regression tasks is measured by misclassification rate and RMSE, respectively. Point-wise standard errors, derived from smoothing spline models \citep{hastie2009elements}, are also depicted to illustrate the variation in error.}
\label{fig: statistical-error}
\end{figure}

Figure \ref{fig: statistical-error} underscores the contribution of Syn in bolstering CatBoost's efficacy in classification and regression. While there is a minor boost in ``FB comments'', the extent of improvement via Syn-Boost is diverse. Classification and regression enhancements span 0.6\% to 17.4\% and 0.03\% to 12.3\%, respectively, against CatBoost's baseline. The magnitude of these enhancements varies by scenario. For instance, datasets like ``Gesture'' and ``House'', which have a larger predictor count, show more significant leaps. In contrast, datasets like ``Adult'' and ``California'' with fewer predictors demonstrate modest gains.

Figure \ref{fig: statistical-error} also highlights the generational effect as the size of synthetic data increases.  Accuracy gains plateau after reaching the estimated reflection point $\hat m$, an estimated optimal size of synthetic data. 
This point represents the peak of statistical accuracy and is consistently greater than raw sample sizes, often by at least five-fold. In scenarios like ``Gesture'', ``Adult'', ``California'', ``House'', and ``Insurance'', Syn-Boost surpasses CatBoost even when raw and synthetic data sizes are equal ($m=n$). This observation suggests that 
the efficacy of Syn-Boost is rooted in the increased sample size $m$, as evidenced by the ``California'' 
and ``House'' datasets that utilized an optimized synthetic-to-raw data ratio of 25:1 or higher. Typically, error curves form a U-shape around a moderate $\hat m$ but shift to an L-shape when $\hat m$ is exceptionally large.

In a supervised setting, Syn-Boost, which employs CatBoost on synthetic data, typically outperforms FNN, except in the ``FB comments'' and ``Abalone'' datasets. When the generation errors are modest, the performance of Syn-Boost over FNN spans from 11.1\% to 14.6\% in classification and 7.2\% to 16.3\% in regression. The reduced performance on the ``FB comments'' and ``Abalone'' datasets, with a decline ranging from 1.4\% to 6.6\%, is chiefly attributed to non-negligible generation errors from the pre-trained TDM \citep{kotelnikov2023tabddpm}. In both scenarios, Syn-Boost may not surpass CatBoost when $m=n$. A similar phenomenon also occurs for other models including decision trees, random forests, and logistic regression without knowledge transfer \citep{kotelnikov2023tabddpm}. We speculate that the generation error in the ``FB comments'' dataset arises from the model architecture's inability to handle large pre-training instances, while the ``Abalone'' dataset's underperformance could be due to insufficient pre-training size. Notably, while FNN outperforms CatBoost in the ``Abalone'', ``FB comments'', and ``Gesture'' datasets, it lags in others. These findings highlight Syn-Boost as a strong competitor against well-established predictive models.

\vspace{3pt}

\noindent\textbf{Knowledge Transfer Across Distinct Distributions.} 
The Adult dataset \citep{adult}, derived from the $1994$ census, includes data on $32,650$ males and $16,192$ females, featuring six numerical and eight nominal attributes such as age, work class, education years, marital status, weekly working hours, and native country. Our goal is to predict whether an adult female's income surpasses \$50K annually using a generative model pre-trained on male data to facilitate knowledge transfer, despite their differing distributions but capitalizing on their similarities.

\begin{figure}[H]
\centering
  \includegraphics[width=\textwidth]{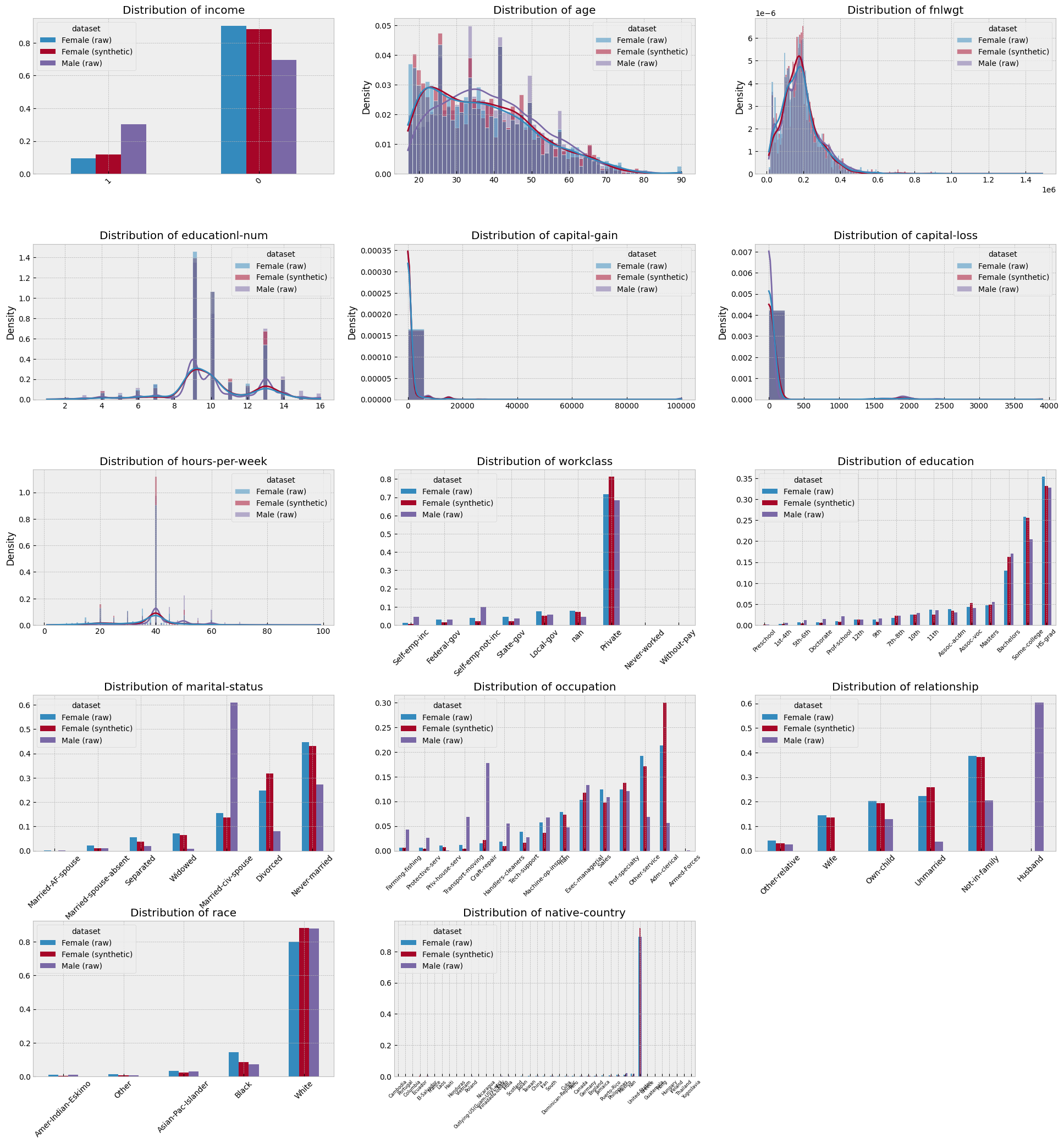}
\caption{Marginal distributions of datasets categorized as female, synthetic female, and male are illustrated, with legends arranged from top to bottom. Normalized bar and kernel density plots represent categorical and numerical features, respectively.}
\label{fig: dist_diff_gender}
\end{figure}

As Figure \ref{fig: dist_diff_gender} depicts, the pronounced differences across genders exist in the distributions of categories like income, age, marital status, occupation, and relationship. A pertinent question arises: can adult male data augment the synthetic data generation and subsequent classification task on adult female data?
To harness this knowledge transfer, we pre-train a TDM solely on the adult-male data with $32,650$ observations. 
We utilize a subset of size $1,350$ from the adult-female data as our raw data, and another independent subset of the same size as the test data for evaluation purposes.

\begin{table}
\centering
\caption{Distribution distance comparisons using FID-scores,  1- and 2-Wasserstein distances between the actual female sample and various synthetic samples. The terms "raw," "pre-trained," and "fine-tuned" denote the raw data, synthetic data from a pre-trained model, and synthetic data from a fine-tuned pre-trained model, respectively. The evaluation sample size for "Female (raw)" is $1,350$, while for "Male (raw)," "Male (pre-trained)," "Female (pre-trained)," and "Female (fine-tuned)," is $32,650$.}
\label{tab:distribution-distance}
\begin{tabular}{lccc}
\hline
                                       & FID (Gaussian) & 1-Wasserstein & 2-Wasserstein \\ \hline
Female (raw) vs Male (raw)           & 1.971          & 1.968         & 2.125         \\
Female (raw) vs Male (pre-trained)   & 2.051          & 1.967         & 2.127         \\
Female (raw) vs Female (pre-trained) & 0.457          & 1.292         & 1.536         \\
Female (raw) vs Female (fine-tuned)  & \textbf{0.249}          & \textbf{1.170}         & \textbf{1.399}         \\ \hline
\end{tabular}%
\end{table}

\begin{figure}
\begin{center}
  \includegraphics[width=\textwidth]{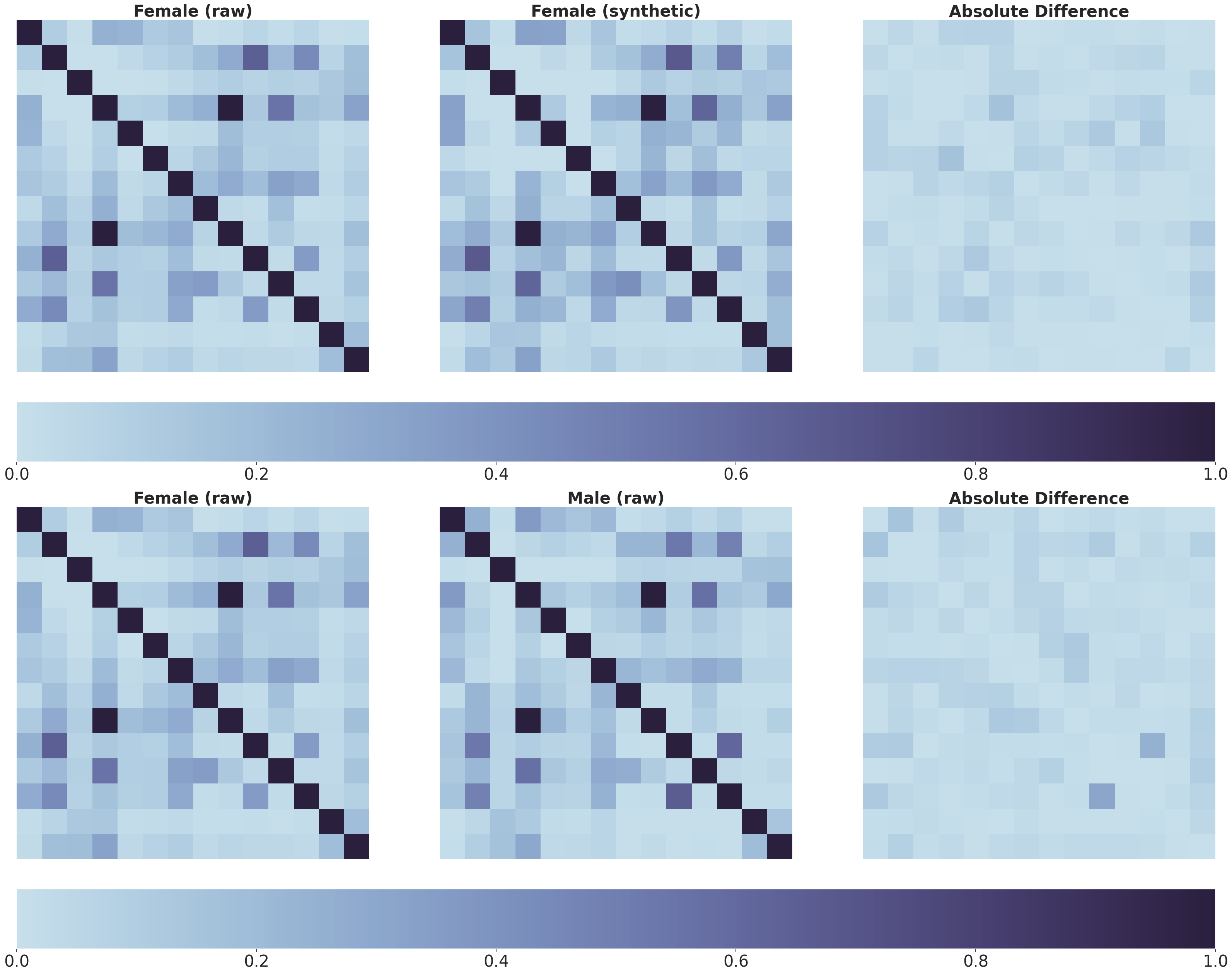}
\caption{Pairwise correlation plots between raw and synthetic female datasets compare those between raw female and male datasets, accompanied by their differences. Dark cells in the difference plots signify pronounced deviations from the female distribution. Pearson's correlation, Correlation Ratio, and Theil's U measure continuous-continuous, categorical-continuous, and categorical-categorical correlations.}
\label{fig: corr_audlt_gender}
\end{center}
\end{figure}

Figures \ref{fig: dist_diff_gender} and \ref{fig: corr_audlt_gender} demonstrate that the synthetic female data produced by TDM aligns more with the adult-female dataset than with the adult-male dataset. This alignment is evident by the Fr\'echet Inception Distance (FID), which measures the distributional differences between the generated and raw data vectors under the Gaussian assumption. The 1- and 2-Wasserstein distances between the empirical distributions of the two datasets provide further evidence (note that FID is 2-Wasserstein distance under Gaussian assumption).
Notably, as shown in Table \ref{tab:distribution-distance}, the male-focused pre-trained TDM, once fine-tuned with a smaller female dataset, crafts synthetic female data with a diminished margin of error compared to models trained solely on the adult-male data and adult-female data. This result confirms that leveraging pre-trained adult-male data with a somewhat distinct distribution can enhance the TDM's generation precision for females. This empirical validation emphasizes the imperative of refining a pre-trained model to maximize knowledge transfer and achieve unparalleled accuracy.

\begin{figure}
    \centering
    \includegraphics[width=\textwidth]{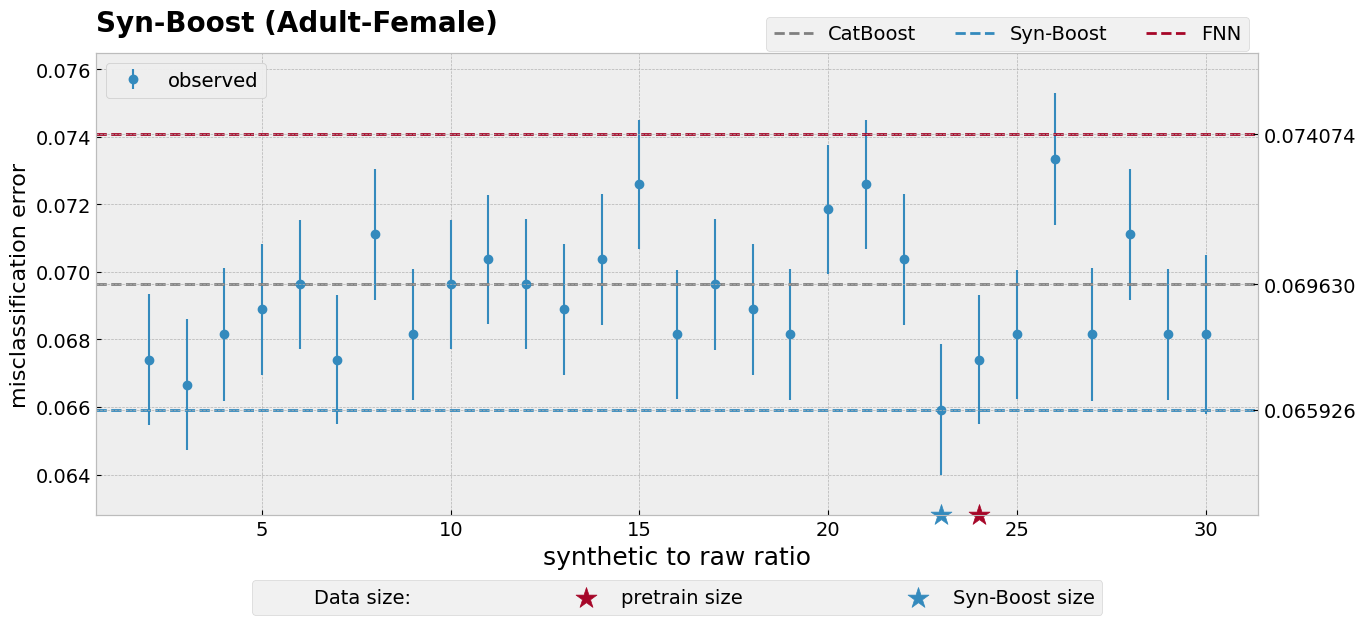}
\caption{Comparative error analysis of CatBoost, Syn-Boost, and FNN, with Syn-Boost and FNN utilizing transfer learning with distinct distributions on the Adult-Female dataset \citep{adult}, with Adult-Male data serving as pre-training data, across various synthetic-to-raw data ratios. Stars indicate the pre-training data size and the tuned sample size for Syn-Boost. The vertical bars, calculated using smoothing spline models \citep{hastie2009elements}, represent the pointwise standard error.}
    \label{fig: statistical-error-female}
\end{figure}

The result of Syn-Boost prediction is illustrated in Figure \ref{fig: statistical-error-female}.
We observe that simply increasing the synthetic size does not necessarily improve the prediction performance, as depicted at the beginning of the Syn-Boost tuning.
The optimal empirical ratio is obtained at $m / n = 23$, which is about the same size as the pre-training adult-male data.
Moreover, Syn-Boost achieves $5.32\%$ improvement compared to CatBoost in misclassification error, and $11.00\%$ improvement compared to the FNN approach, indicating the potential of this method.

Both case studies highlight the effectiveness of the Syn framework in enhancing statistical accuracy through synthetic data generation.
Syn's robust performance primarily stems from the generative capabilities of diffusion models coupled with the application of knowledge transfer. These elements enhance the generative model's generation accuracy by accurately estimating the distribution of raw data over low-dimensional manifolds \citep{oko2023diffusion}. However, it is crucial to recognize that the Syn framework's success also hinges on thoughtful modeling and predictor selection.
The case studies also explain the results of \citet{kotelnikov2023tabddpm} regarding the potential pitfalls in a prediction task when employing synthetic data to train machine learning models, a concern mentioned in the Introduction. Low-fidelity synthetic data, resulting from substantial generation errors, can negatively impact statistical accuracy. The study suggests that employing knowledge transfer from relevant studies is a strategy to reduce generation errors.

\subsubsection{Simulation}\label{sec: simulation-prediction}

To investigate how generation errors impact the efficacy of Syn-Boost, we conducted simulations with access to ground truth data. We consider a regression model:
\begin{equation}\label{eq: simulation-prediction}
\mb Y = 8 + \mb X_1^2 + \mb X_2 \mb X_3 + \cos(\mb X_4) + \exp(\mb X_5 \mb X_6) + 0.1 \mb X_7 + \mb \epsilon,
\end{equation}
where $\bm X = (\bm X_1, \ldots, \bm X_7)$ is uniformly distributed over $[0,1]^7$ ($\text{Uniform}(0, 1)^7$) and $\mb \epsilon$ follows a normal distribution with mean zero and standard deviation $0.2$, $N(0, .2^2)$. In  \eqref{eq: simulation-prediction}, we generate a dataset of 700 samples, dividing it into 500 for training and 200 for validation. To demonstrate the impact of effective versus ineffective generators on downstream tasks, we pre-train tabular diffusion models \citep{kotelnikov2023tabddpm}) with two sizes, $1000$ and $5000$. It is noteworthy that pre-trained models typically use considerably larger training sizes. To evaluate the distribution discrepancy between raw and synthetic samples, we employ the 2-Wasserstein distance, defined as their distributional distance and determined by solving an optimal transport problem using appropriate metrics\footnote{\url{https://pythonot.github.io/quickstart.html\#computing-wasserstein-distance}}, as detailed in Table \ref{tab:sim-prediction} for reference.

We evaluate Syn-Boost's root mean square error (RMSE) on the prediction performance of synthetic data generated from a pre-trained model, both with and without fine-tuning on raw training data. These scenarios represent the outcomes with ineffective and effective generators, respectively. For comparative purposes, we also assess the RMSE of CatBoost, trained on raw data, and provide the square root of Bayes error, which is 0.2 by design.

\begin{figure}
\includegraphics[width=\textwidth]{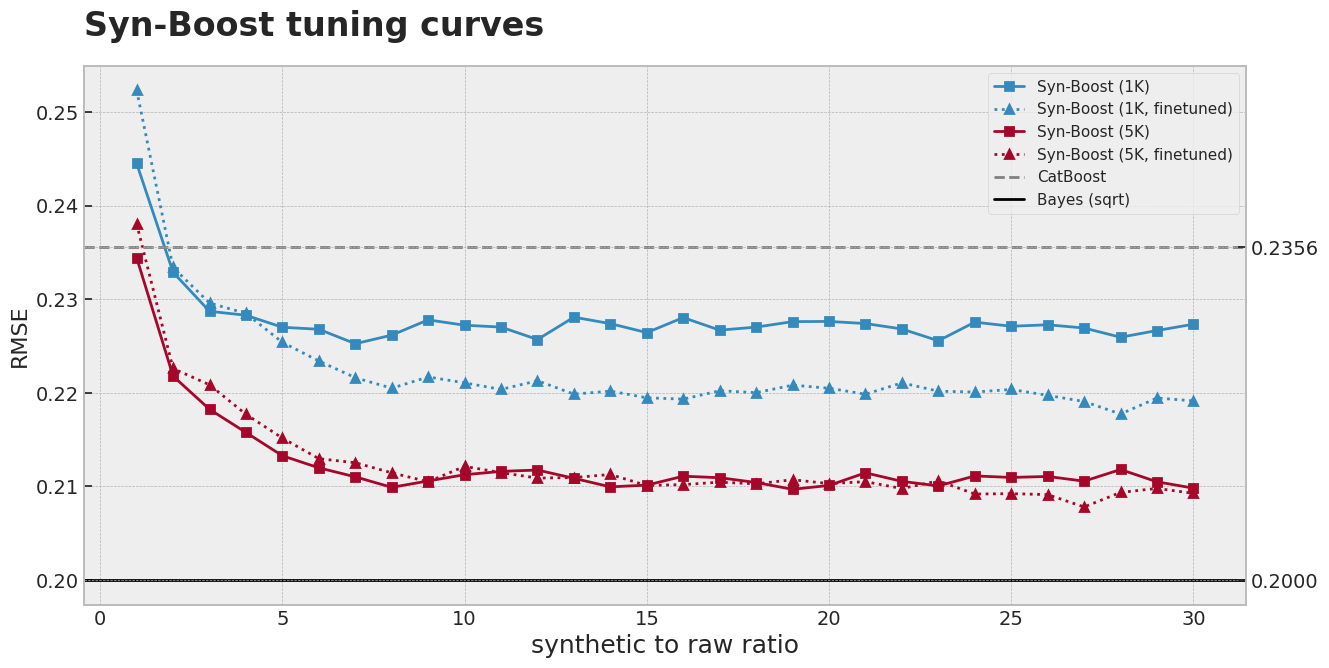}
 \caption{RMSEs of Syn-Boost as a function of the synthetic to raw data ratio in simulations described in \eqref{eq: simulation-prediction}. It contrasts scenarios with small (pre-training size of 1,000) and large (pre-training size of 5,000) generation errors, as well as the generation with (dotted lines) and without (solid lines) fine-tuning on raw data. The grey dashed lines denote the RMSE of CatBoost, while the black solid lines indicate the square root of the Bayes error.} 
\label{fig: simulation-prediction-finetuned}
\end{figure}

As depicted in Figure \ref{fig: simulation-prediction-finetuned}, Syn-Boost attains an RMSE closer to the Bayes error when employing an effective pre-trained generator, thus outperforming CatBoost trained on raw data. In contrast, with an ineffective pre-trained generator, the RMSE of Syn-Boost is similar to the CatBoost error, far from the Bayes error. In practice, we recommend fine-tuning a pre-trained model rather than using it directly. As illustrated in Figure \ref{fig: simulation-prediction-finetuned},   fine-tuning pre-trained generators can further improve the performance of Syn-Boost. Table \ref{tab:sim-prediction} supports the observed generation effects, which supplements Figure \ref{fig: simulation-prediction-finetuned}.

\begin{table}
\centering
\caption{The impact of generation error on Syn-Boost's RMSE in regression. The generation error is quantified using the 2-Wasserstein distance, comparing a synthetic sample of size $50,000$ with an independent sample of the same size from 
the raw distribution in \eqref{eq: simulation-prediction}. The terms "CatBoost," "Syn-Boost," and "Bayes" represent the RMSE for CatBoost applied to raw data, CatBoost applied to synthetic data, and the Bayesian error (in square root), respectively. "Syn-Raw Ratio" indicates the ratio of synthetic to raw data.}
\label{tab:sim-prediction}
\begin{tabular}{@{}ccccccc@{}}
\toprule
Pre-training size & Fine-tuning & 2-Wasserstein & Syn-Raw ratio & Syn-Boost & CatBoost              & Bayes             \\ \midrule
1000              & No          & 0.340         & 7               & 0.220          & \multirow{4}{*}{0.236} & \multirow{4}{*}{0.200} \\
1000              & Yes         & 0.259         & 28              & 0.211          &                        &                        \\
5000              & No          & 0.241         & 19              & 0.206          &                        &                        \\
5000              & Yes         & 0.233         & 27              & 0.203          &                        &                        \\ \bottomrule
\end{tabular}%
\end{table}

\subsection{Feature Relevance for Tabular Data} \label{feature}

This subsection concerns testing the relevance of features for predicting the outcome of the response variable $Y$ by
a machine learner using a candidate feature vector $\mb X$.

Define the subvector $\X_S$ by $\X_S = (X_j: j \in S)$, where $S$ is a subset of the features. Our objective is 
significance testing of $\X_S$ in its functional relevance to $Y$.
To assess the influence of $\X_S$, we use the differenced risk $R(f^*)- R(f_{S^{c}}^*)$. Here, $f_{S^c}=f(\X_{S^c})$, and 
$f^*$ and $f^*_{S^{c}}$ are the optimal prediction functions in the population, defined as $f^*=\argmin_{f} R(f)$ and $f^*_{S^c}=\argmin_{f_{S^c}} R(f_{S^c})$. The risks, $R(f)$ and $R(f_{S^c})$, are given by $R(f) = 
\E\big(L(f(\X), Y) \big)$ and 
$R(f_{S^c}) = \E\big(L(f(\X_{S^c}), Y) \big)$, where $\E$ represents the expectation. Now, we introduce the null and its alternative hypotheses $H_0$ and $H_a$:
\begin{equation} \label{eqn:loss_testing} H_0: R(f^*)- R(f^*_{S^c})=0, \ \text{versus} \ H_a: R(f^*) - R(f^*_{S^c})<0. \end{equation}
Rejecting $H_0$ at a significance level implies feature relevance of $\X_S$ for predicting $Y$. 
It is worth mentioning
that we target the population-level functions $f^*$ and $f^*_{S^c}$ in \eqref{eqn:loss_testing}.

In \eqref{eqn:loss_testing}, \cite{dai2024significance} developed an asymptotic test tailored to black-box learning models. Building upon this foundation, we illustrate how Syn-Test can bolster the power of this traditional test on raw samples by enlarging the synthetic data size while circumventing the necessity to derive the asymptotic distribution of a test statistic.

For Syn-Test, we adhere to Steps 1-4, as delineated in Section \ref{sec: hypothesis-testing}, to examine the relevance of feature set $\X_S$ to outcome $Y$, employing CatBoost \citep{dorogush2018catboost} as the learning algorithm. Here, we engage a diffusion model, TDM \citep{kotelnikov2023tabddpm}, to generate synthetic data. Initially, we adapt the original test statistic from \citep{dai2024significance} to suit synthetic data as follows:

\begin{eqnarray} \label{tests} T = \frac{R_m(\tilde{f}_{S^c}) - R_m(\tilde{f})}{SE(R_m(\tilde{f}_{S^c}) - R_m(\tilde{f}))},
\end{eqnarray}
where $R_m(\cdot)$ denotes the empirical risk, evaluated on an inference sample $\tilde \Z_1^{(m,d)}$ in Step 1 and $\tilde \Z_1^{(m)}$ with $m=\hat m$ in Step 3 of Syn-Test, $\tilde f$ and $\tilde{f}_{S^c}$ denote 
the estimated predictive function function with and without $S$, and $SE$ denotes the standard error. 
Note that a large value of $T$ indicates the relevance of tested features to the response.


In \eqref{tests}, we reject $H_0$ if $T$ manifests 
as large.  To compute the test statistic values in Steps 1 and 3, we generate an additional synthetic sample of size $2N$ 
and split it evenly into two subsamples. Using the first subsample, we train
a CatBoost model $\E (Y | \X)=f(\X)$ to forecast $Y$ employing all features $\X$, resulting in $\tilde f$. In parallel, we train another CatBoost model $\E (Y | \X_{S^c})=f(\X_{S^c})$ using features $\X_{S^c}$, yielding $\tilde f_{S^c}$.
By employing the synthetic sample to compute full and null predictive models, yielding $\tilde f$ and $\tilde f_{S^c}$, we can mitigate the intrinsic bias and asymptotics highlighted in \citep{dai2024significance} stemming from a limited 
inference size. This behavior is demonstrated in Figure \ref{fig: syninf_allinone} (middle row).

To refine a generative model under the null hypothesis, researchers often employ permutation by replacing redundant predictor vectors $\X_S$ with irrelevant values \citep{dai2024significance}. However, this approach may not preserve the correlation structures between $\bm X_S$ and $\X_{S^c}$. Addressing this issue, we introduce a novel method that maintains these correlation structures while ensuring the feature irrelevance of $\X_S$ on $Y$ given the rest of the features.
Our procedure involves two steps:
\begin{enumerate}
\item We first train a predictive model to estimate $\E(\X_S | \X_{S^c}) = g(\X_{S^c})$.

\item We generate synthetic data tuples $(Y, \X_S, \X_{S^c})$ using this model. Then, we modify these tuples  by replacing $\X_S$  with the predicted values $g(\X_{S^c})$, resulting in new tuples $(Y, g(\X_{S^c}), \X_{S^c})$. This process ensures compliance with a specific subclass under the risk invariance of $H_0$ by creating conditionally independent tuples, as described by \citet{dai2024significance}.
\end{enumerate}

\noindent Consequently, we obtain modified tuples $\tilde{\bm Z}^{(m)}=(\tilde Y_i,\tilde{\bm X}_{iS}, \tilde{\bm X}_{iS^c})_{i=1}^m$, which adhere to the feature irrelevance hypothesis $H_0$.

Finally, we designate an MC size of $D=1,000$ for estimating both the null distribution and the Type-I error in Step 1 of
Syn-Test. The parameters are set as $\alpha = 0.05$, $\varepsilon = 0.01$, and the optimal $\hat m$ will be tuned based on the ratios $m/n \in \{1, 2, \ldots, 20\}$.


\subsubsection{Real-Benchmark Examples} \label{sec: inference-real}

Knowledge transfer profoundly impacts the behavior of synthetic data, affecting critical downstream tasks, including inference. To illuminate this relationship, we employ Syn-Test to assess feature relevance using the gradient boosting method, CatBoost \citep{dorogush2018catboost}. We explore this in a regression context with the California dataset \citep{california} and a classification context using the Adult dataset \citep{adult}, maintaining the experimental setup detailed in Section \ref{prediction}. Within these frameworks, we examine the influence of knowledge transfer on synthetic data generation. Concurrently, we evaluate the efficacy of Syn-Test in identical scenarios and those that are distinct yet closely related.

To contrast Syn-Test with its traditional counterpart, consider the significance test in \eqref{tests}. When the finite-sample null distribution of $T$ is unknown, the asymptotic distribution of the test statistic may require stringent assumptions \citep{dai2024significance}. To circumvent this, we use synthetic samples generated from TDM to approximate the null distribution, as in \citep{liu2023perturbation}. Contrary to that approach, we refrain from using data perturbation, thus eliminating the requirement to maintain the rank property for privacy protection.

\vspace{3pt}
\noindent\textbf{Knowledge Transfer from Identical Distributions.} Drawing from the 1990 U.S. census, the California dataset offers a glimpse into median house values through eight specific attributes. These include the longitude and latitude of the property, its median age, the total room count, bedroom count, block population, household count within the block, and the median household income.

\begin{figure}
    \centering
    \includegraphics[width=\textwidth]{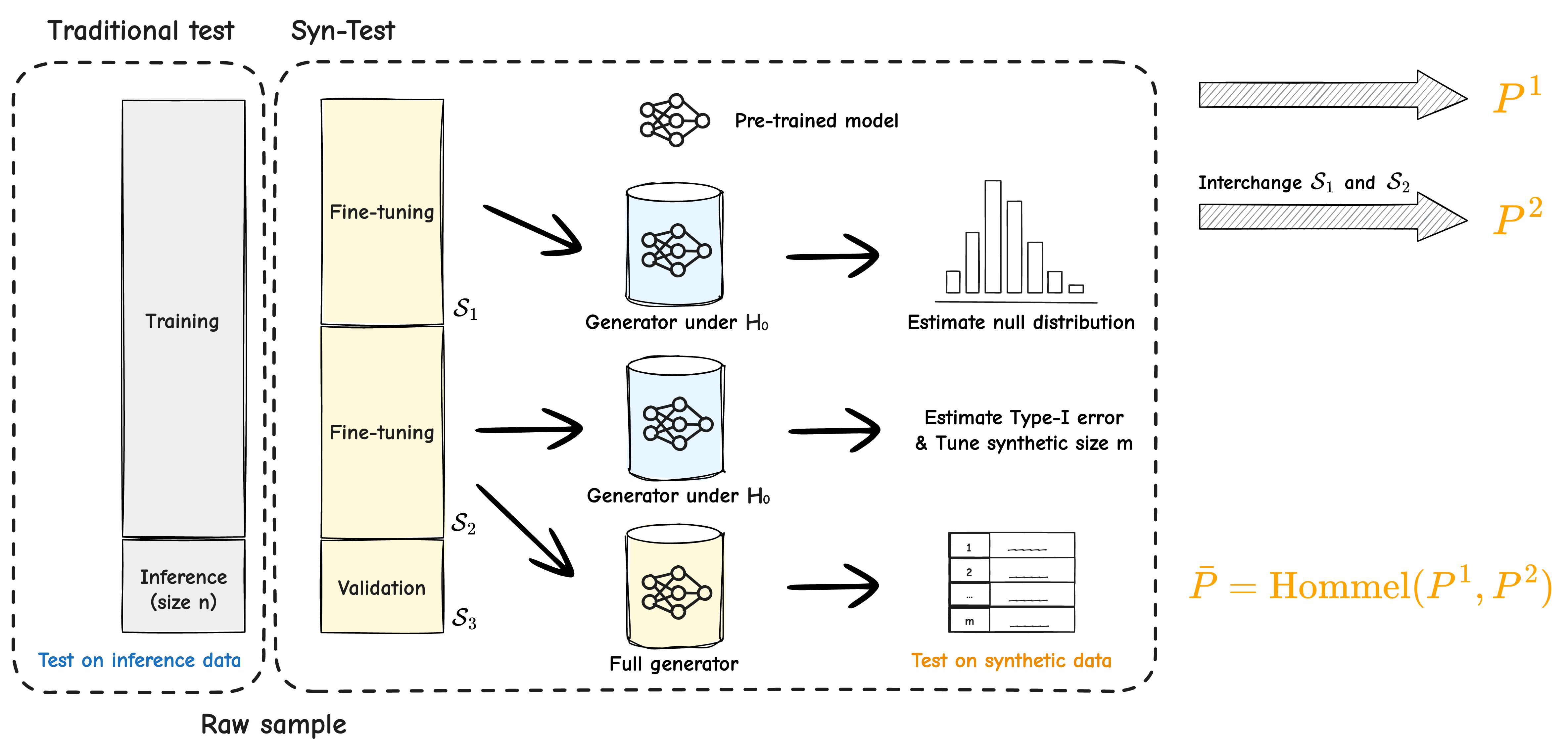}
    \caption{Illustration of Syn-Test in comparison with traditional feature significance test for black-box models. For the traditional approach, one needs a training split to train the black box model independently \citep{dai2024significance}, which we split evenly into two subsamples for fine-tuning. More details can be found in Section \ref{sec: hypothesis-testing}.}
    \label{fig:syntest}
\end{figure}

To facilitate knowledge transfer, we initially pre-train a TDM using the pre-training split with $13,209$ observations. For significance testing in \eqref{eqn:loss_testing}, we adapt the one-split black-box test statistic in \citep{dai2024significance} with a training sample of size $6,605$ and an inference sample of size $826$.
To perform the Syn-Test, we follow the splitting scheme illustrated in Figure \ref{fig:syntest}.
In detail, we divide the training sample equally into two subsets, $\mc S_1$ and $\mc S_2$, for fine-tuning purposes. Additionally, we utilize $\mc S_3$ as a validation sample for both model training and fine-tuning.
More details on the Syn-Test method can be found in Section \ref{sec: hypothesis-testing}.

\begin{figure}
\centering 
\includegraphics[width=\textwidth]{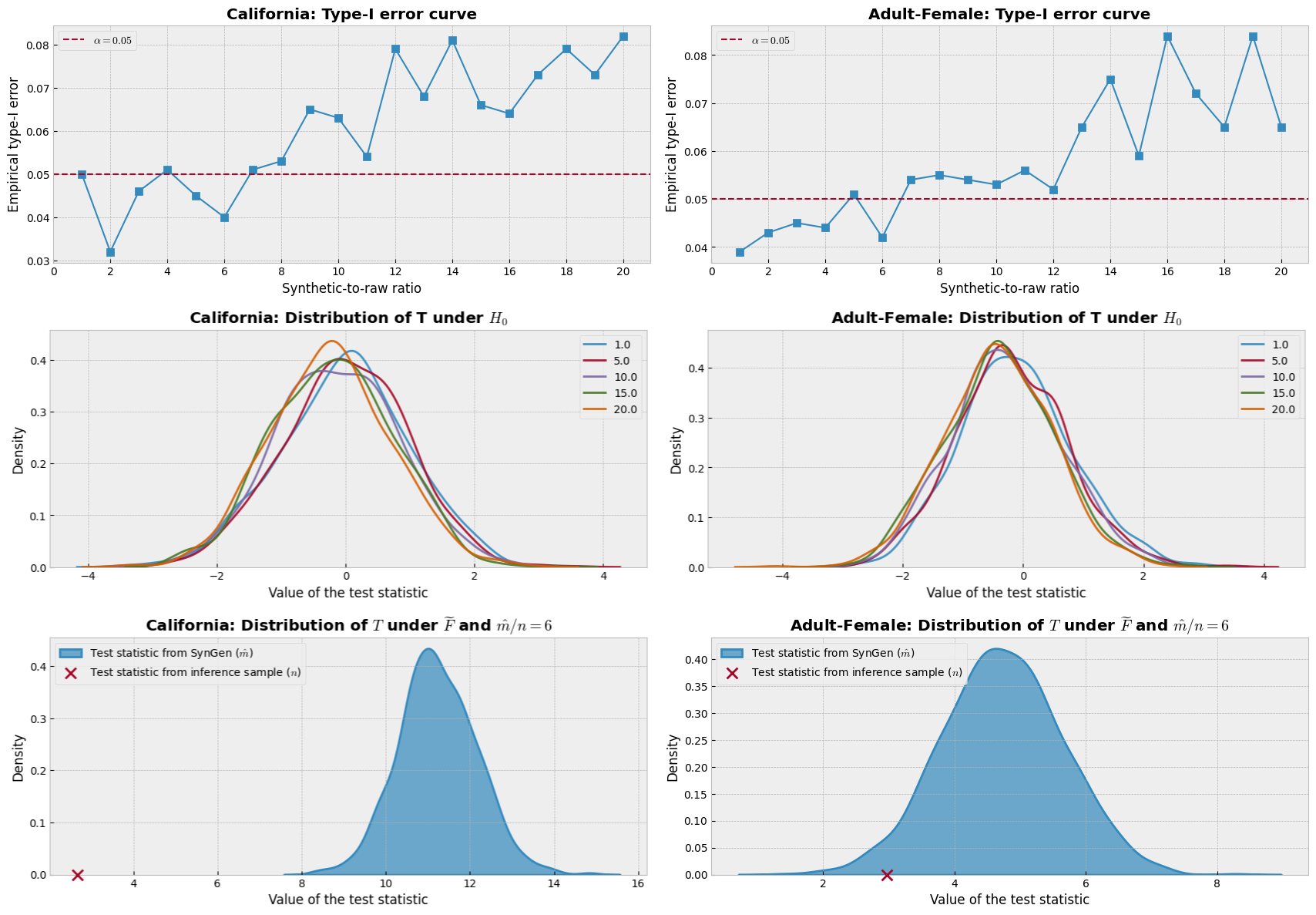}
\caption{Evaluation of Syn-Test for feature relevance using CatBoost, with parameters $\alpha = 0.05$, $\varepsilon = 0.01$, and $D=1000$. Left: Analysis on California dataset \citep{california} using a sample of $n=826$, focuses on the feature \textit{MedInc} for predicting \textit{MedHouseVal}. Right: Study on Adult dataset \citep{adult} with $n=300$, examines features \textit{Age, Education-num, Hours-per-week} for classifying \textit{income} over 50K. Top: Type-I error curves display Syn-Test errors (blue) and the threshold of $\alpha=0.05$ (red dashed). Middle: Null distributions, shown with five synthetic-to-raw data ratios (1,5,10,15,20), suggest small bias in estimating 
$f^*$ and $f^*_{S^c}$. Bottom: Power analysis compares the Syn-Test statistic distribution (blue kernel, size $\hat m$) with the traditional test's statistic value (red cross, size $n$) under alternative hypothesis $H_a$, highlighting greater power.}
\label{fig: syninf_allinone}
\end{figure}

As depicted in Figure \ref{fig: syninf_allinone} (top left), the empirical Type-I error initially descends, then rises, ultimately surpassing the $\alpha=0.05$ level as the ratio of synthetic-to-raw size $\hat m/n$ grows. The estimated maximum ratio $\hat m/n=6$ preserves the Type-I error control. In other words, we can augment the sample size to sixfold the raw sample size $n$. This observation aligns with the generational effect we observed in predictive modeling for classification and regression. Consequently, $\hat m= 6n$ notably enhances the power of Syn-Test, as indicated in 
Figure \ref{fig: syninf_allinone} (bottom left), where
the test statistic distribution shifts to the right, increasing the power
to reject $H_0$ when it is false.

\vspace{3pt}
\noindent\textbf{Knowledge Transfer Across Distinct Distributions.}
In the study of the Adult dataset \citep{adult}, we conduct a significance test to assess the relevance of three features—age, education years, and weekly working hours—in predicting whether an adult female's income exceeds \$50K annually.

To enhance the test's power for females, we generate synthetic data for females through knowledge transfer from the male dataset. This methodology aligns with the approach in the latter part of Section \ref{prediction-real}. Here, we pre-train a TDM on a dataset comprising $32,650$ adult male records. For our hypothesis testing in \eqref{eqn:loss_testing} focused on adult females, we use a training set of $2,700$, split equally between $\mc S_1$ and $\mc S_2$. Additionally, we use an inference sample of $300$ that also serves as a validation set $\mc S_3$ in the Syn-Test. This approach, as outlined in Section \ref{sec: hypothesis-testing} and illustrated in Figure \ref{fig:syntest}, involves fine-tuning the pre-trained TDM with $\mc S_1$ and $\mc S_2$ while utilizing $\mc S_3$ to prevent overfitting during model training and refinement. Table \ref{tab:distribution-distance} and Figures \ref{fig: dist_diff_gender} and \ref{fig: corr_audlt_gender} show the effectiveness of the fine-tuned TDM and address the importance of knowledge transfer for reducing generation error.

Figure \ref{fig: syninf_allinone} (right column) highlights two key observations. First, the top figure demonstrates that $\hat m = 6n$ consistently controls the Type-I error estimated from Syn-Test.
Second, the bottom plot exhibits a pronounced rightward shift in the test statistic distribution when comparing synthetic data to those derived from raw data. This shift indicates an enhancement in the test's power, directly attributable to the augmentation of the sample size.

\subsubsection{Simulation} \label{sec: inference-simulation}

In this section, we evaluate the effectiveness of Syn-Test in controlling Type-I errors through simulation studies. We use the model in  \eqref{eq: simulation-prediction} with a modification: an additional feature, $\mb X_8$, is included but does not contribute to the model. Here, $\mb X = (\mb X_1, \ldots, \mb X_7, \mb X_8)$ is distributed uniformly on $[0,1]^8$  ($\text{Uniform}(0, 1)^8$), and $\mb \epsilon$ follows a normal distribution $N(0, 0.2^2)$. We assess the relevance of the feature $\mb X_8$ in prediction, thereby examining the control capability of Type-I error by Syn-Test. 
We will use the one-split test statistic proposed by \citep{dai2024significance} in Syn-Test, although we don't rely on their asymptotic theory for the test statistic.

For our experiments, we split a raw sample into a training set and an inference set with $1000$ and $200$ samples, respectively. 
Additionally, we use a distinct pre-training sample of $10000$ to train the TDM on $(\mb Y, \mb X)$. This pre-trained model is subsequently fine-tuned on the raw training set according to the Syn-Test's procedure. The inference sample, consisting of 200 instances, is utilized to validate the training of $\tilde f$ and $\tilde f_{S^c}$ for the test statistic in \eqref{tests}.

Moreover, we employ an MC size of $D = 1000$ with parameters $\alpha = 0.05$ and $\epsilon = 0.01$, and explore synthetic-to-raw ratios from ${1, 2, \ldots, 20}$ to fine-tune the synthetic size $m$.

\begin{figure}
\begin{center}
  \includegraphics[width=\textwidth]{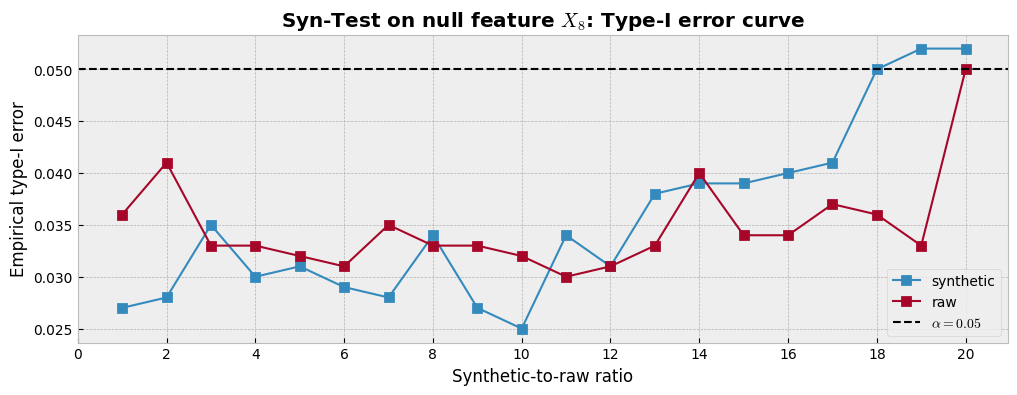}
\caption{Tuning curves of Syn-Test using synthetic data versus the same test using independent raw data for selecting the synthetic size $\hat m$ in testing the null feature $\mb X_8$ with 
$\alpha = 0.05$ and $\epsilon = 0.01$.}
\label{fig: sim_inf_tuning}
\end{center}
\end{figure}

As shown in Figure \ref{fig: sim_inf_tuning}, the tuning curve of Syn-Test with synthetic data demonstrates a comparable performance to that of the same test when using independent raw data, particularly in terms of generational effect, 
while exhibiting similar patterns of variation.
This finding indicates that Syn-Test effectively controls the Type-I error with synthetic data, aligning with observations from raw data. Finally, we adopt the conservative approach in selecting $m$, choosing $\hat m / n = 18$, in accordance with the guidelines 
in Section \ref{sec: hypothesis-testing}.

\begin{figure}
\begin{center}
  \includegraphics[width=\textwidth]{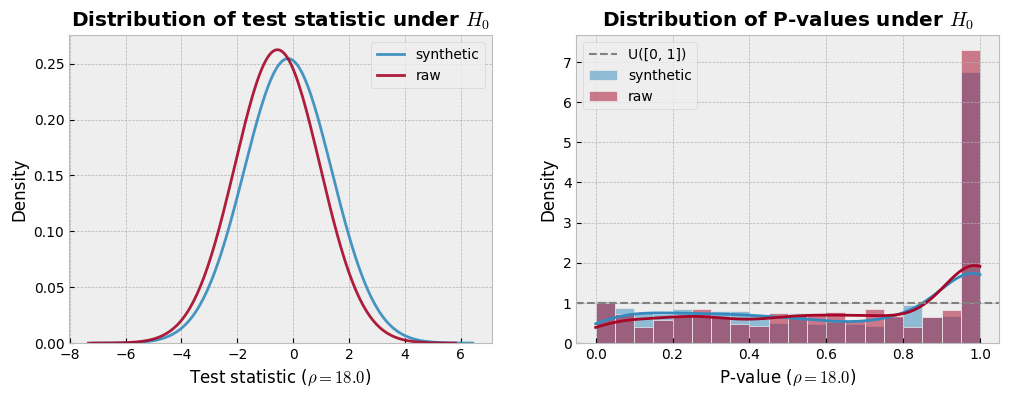}
\caption{Comparative distributions of the test statistic (left) and P-values (right) under $H_0$ for testing $\mb X_8$ using
Syn-Test with synthetic data versus raw data. Note that the P-value distributions approximate the uniform distribution $U[0,1]$,
apart from the point mass at $1$ due to Hommel's combination technique.}
\label{fig: sim_inf_control}
\end{center}
\end{figure}

As illustrated in Figure \ref{fig: sim_inf_control}, the estimated null distribution curve derived from synthetic data with $\hat m / n = 18$ closely resembles that based on independent raw data, albeit with a slight shift. This observation suggests minor generation errors by our generators. Furthermore, the distribution of P-values from Syn-Test using synthetic data under the null hypothesis $H_0$ with $\hat m / n = 18$ aligns well with that based on raw data. These plots demonstrate the effectiveness of Syn-Test in controlling Type-I error.

\section{Data Privacy}
\label{privacy}                             

The Syn framework can address privacy concerns using synthetic data generated by generative models trained to mimic the distribution of raw data. Unlike raw data, synthetic data imposes fewer privacy risks. However, it's not entirely immune to reverse engineering attacks. These vulnerabilities mainly stem from the model's parameters, which attackers could potentially infer from the synthetic data. In real-world applications like healthcare and finance, where data sensitivity is paramount, ensuring robust privacy measures is crucial.

To bolster data privacy, one may consider a privacy protection standard known as $(\varepsilon,\delta)$-differential privacy \citep{dwork2006our}, recognized as a gold standard in data privacy. Notably, it was implemented in the 2020 U.S. decennial census, demonstrating its practical applicability and effectiveness. This approach effectively safeguards against various privacy threats, including reverse engineering, re-identification, and inference attacks.          

The definition of $(\varepsilon,\delta)$-differential privacy considers an adjacent realization $\z'$ differing from
a realization of an original sample $\Z=(\bm Z_i)_{i=1}^n$ by just one observation. It revolves around a privatization
mechanism $\M$, mapping from original dataset $\Z=(\Z_i)_{i=1}^n$ into a privatized version $\tilde{\Z}^{(m)}=(\tilde \Z_i)_{i=1}^m$. For $\M$ to be $(\varepsilon,\delta)$-differential private \citep{dwork2006our}, it satisfies: For any small $\varepsilon \geq 0$ and $\delta>0$ and any measurable set $B$:
\begin{eqnarray} \label{diff}
P\big(\M(\Z) \in B|\Z=\z\big) \leq e^{\varepsilon} P\big(\M(\Z)
\in B|\Z=\z'\big)+\delta,                   
\end{eqnarray}
where $\varepsilon>0$ is the privacy budget, controlling the strength of privacy protection. Smaller $\varepsilon$ values indicate stronger privacy, while $\delta$ is a small probability allowance for the privacy guarantee, acknowledging minimal inherent risk. This definition accommodates $\varepsilon$-differential privacy with $\delta=0$.

To generate synthetic data satisfying \eqref{diff}, we may employ techniques like Differentially Private Stochastic Gradient Descent (DP-SGD). DP-SGD injects a certain amount of Gaussian noise into the gradient updates during model training,
ensuring that the model satisfies the desired privacy guarantees \citep{abadi2016deep}.
Differentially-private diffusion models \citep{ghalebikesabi2023differentiall} is an example of a point. Further exploration is warranted.

Unlike traditional differentially private samples, differentially private synthetic variants through these methods preserve the distributional characteristics of the original data. This preservation is advantageous as it can boost statistical analysis through sample size argumentation while enhancing privacy. However, this approach often requires more extensive model training due to the added noise, which can be a computational challenge.


\section{Discussion: Future of Data Science} \label{future}

This article unveils the Syn paradigm—a novel approach to data analytics using high-fidelity synthetic 
data derived from real-world insights. By addressing challenges in traditional data analytics, such as 
data scarcity and privacy issues, this paradigm underscores that high-fidelity synthetic data can amplify the precision and efficiency of data analytics with sample size augmentation,  as evidenced by the case studies herein. However, it is crucial to acknowledge the generational effect inherent to the Syn framework. Consequently, a fusion of raw and synthetic data 
is essential for unlocking their potential, offering fresh perspectives for both scientific and engineering domains.

In this study, although our primary focus is on tabular data—the predominant form in statistical analysis—we emphasize that the Syn framework's utility extends across disparate types of data given available tools such as multimodal diffusion models \citep{xu2023versatile}.  Its utility is reinforced by recent progress in generative modeling for vision, natural language, and multimodal data processing. This adaptability allows the Syn framework to perform comprehensive analyses of complex, multimodal data, such as electronic health records, which typically include a mix of standard tabular data, image scans, and unstructured text.

The importance of synthetic data in driving progress in data science and AI is increasingly recognized. However,  a comprehensive evaluation of the Syn framework across a broad spectrum of applications is imperative.  One key strategy for assessing the synthetic sample's fidelity or reliability is comparing it with an independent validation sample that reflects the original data's distribution. This assessment can be conducted through (i) exploratory analysis, like histograms and correlation matrices for visual inspection, and (ii) distributional distances, such as the Wasserstein distance, for a quantitative evaluation. Furthermore, in supervised learning tasks, the model's performance can be scientifically validated using cross-validation metrics. Overall, validating generative models for more complex applications presents unique challenges, making the exploration of model robustness and data privacy critical areas for future research.

The future of data science may pivot on our capability to harness raw and synthetic data. Large pre-trained generative models,
equipped with extensive knowledge, offer a promising pathway. These frameworks distill domain knowledge, a testament being the successes of Generative Pre-trained Transformers in text and imagery contexts. As developing domain-centric generative models is gaining momentum, these generative models promise significant enhancements in synthetic data generation, paving the way for breakthroughs across a wide range of disciplines.


\begin{appendix}

\section{Proofs} \label{appendix}

\subsection{Proof of Theorem \ref{thm1}}
\begin{proof}
Let $\text{LTV} \doteq f(\text{TV}(\tilde F, F))$.
Let $g(m) = C_{\mb \theta} m^{-\alpha} +
m \text{LTV}$. The assumption that $\text{LTV} > \Er(\widehat{\mb \theta}(\tilde \Z^{(m_0)}))/m^*$ implies
that $g(m^*)-C_{\mb \theta} (m^*)^{-\alpha} > \Er(\widehat{\bm \theta}(\tilde \Z^{(m_0)}))$. Moreover,
note that $\Gr^{(m)} \geq \Gr^{(m^*)}$ if $m > m^*$. Hence, when $m>m^*$,
\begin{align*}
    \Er(\widehat{\bm \theta}(\tilde \Z^{(m)})) 
    &\geq C_{\mb \theta} (m^*)^{-\alpha} + \Gr^{(m)} + C_{\mb \theta} (m^{-\alpha} - (m^*)^{-\alpha})\\
    &\geq C_{\mb \theta} (m^*)^{-\alpha} + \Gr^{(m^*)} + C_{\mb \theta} (m^{-\alpha} - (m^*)^{-\alpha})\\
    &\geq g(m^*) - C_{\mb \theta} (m^*)^{-\alpha} > \Er(\widehat{\bm \theta}(\tilde \Z^{(m_0)})).
\end{align*}
This completes the proof of (ii).  Moreover, by the definition of $m_0$ and (ii), $m_0 \leq m^* < +\infty$.
This completes the proof of (i).
\end{proof}


\subsection{Proof of Theorem \ref{thm2}}
\begin{proof}
By definition, 
\begin{align*}
|\Gr_m| &= \left | \int
L(\widehat{\bm \theta}_m(\bm t),\bm \theta) ~ (d F(\bm t) - d \tilde F(\bm t)) \right |\\
& \leq 2U\text{TV}(F_{\mb Z^{(m)}}, F_{\tilde {\mb Z}^{(m)}}) \leq 2mU\text{TV}(\tilde F, F),
\end{align*}
yielding \eqref{cor_eq_1}. This, together with the assumption that $\Er(\widehat{\bm \theta}(\Z^{(m)})) = C_{\mb \theta} m^{-\alpha}$ 
yields that
$$
\Er(\widehat{\bm \theta}(\tilde \Z^{(m_0)})) \leq \Er(\widehat{\bm \theta}(\tilde \Z^{(s_0)})) \leq\Er(\widehat{\bm \theta}(\tilde \Z^{(m_0)})) \leq \Er(\widehat{\bm \theta}(\tilde \Z^{(s_0)}))
$$ 
when $s_0 = \left (\frac{2U}{\alpha C_{\mb \theta}}\right )^{\frac{\alpha}{1 + \alpha}} \cdot \text{TV}(\tilde F,F)^{-\frac{1}{1 + \alpha}}$, yielding the right hand of \eqref{cor_eq_1}.
This gives an optimal upper bound of $\Er(\widehat{\bm \theta}(\tilde \Z^{(m_0)}))$ in \eqref{cor_eq_2}.
Moreover, \eqref{cor_eq_2} implies that $\Er(\widehat{\bm \theta}(\tilde \Z^{(m_0)})) \leq \Er(\widehat{\bm \theta}(\Z^{(n)}))$ provided that $\text{TV}(\tilde F,F)$ is sufficiently small, sufficiently,
$\text{TV}(\tilde F,F) \leq C_{\mb \theta, \alpha, U} \cdot n ^{-1 - \alpha}$ where $C_{\mb \theta, \alpha, U} = C_{\mb \theta} (2U)^{-1} \left (\alpha^{\frac{-\alpha}{1+ \alpha}} + \alpha^{\frac{1}{1 + \alpha}} \right)^{-\frac{1 + \alpha}{\alpha}}$. 
This completes the proof.
\end{proof}

\subsection{Proof of Theorem \ref{thm3}} 
\begin{proof}
Let $\tilde {\mb Z}$ be a random vector following $\tilde F = F_{\tilde {\mb Z} \mid \mb Z^{(n)}}$ and $\mb Z$ be one following $F$.
First, we bound
the empirical distribution $F^D_{\tilde T}(x)=D^{-1} \sum_{d = 1}^D X^{(m,d)}$ in \eqref{bounds}, where
$X^{(m,d)}=I(T(\tilde \Z^{(m,d)}) \leq x) \in [0, 1]$ for any $x \in \mc R$.
Let $F_{\tilde T}(x)=\E_{\tilde \Z^{(m,1)} \mid \Z^{(n)}} X^{(m,1)}$. Note that $\tilde \Z^{(m,d)}$ is a conditionally 
independent sample  of size $m$ given $\Z^{(n)}$ following $\tilde F=F_{\tilde \Z \mid \Z^{(n)}}$. By Hoeffding's Lemma, $\E_{\tilde \Z^{(m,d)} \mid  \Z^{(n)}} \exp(s(X^{(m,d)} - \E_{\tilde \Z^{(m,d)} 
\mid \Z^{(n)}} X^{(m,d)})) \leq \exp(s^2/8)$ a.s. for any $s>0$, where $\E_{\tilde \Z^{(m,d)} \mid \Z^{(n)}}$ is the conditional expectation with respect to $\tilde \Z^{(m,d)}$ given $\Z^{(n)}$.
By Markov's inequality and the conditional independence between $\tilde \Z^{(m,1)}, \ldots, \tilde \Z^{(m,d)}$ given $\Z^{(n)}$, for any $t>0$ and $s=4 t$,
{\small
\begin{align*}
& \Pr\Big(\abs{F^D_{\tilde T}(x) - F_{\tilde T}(x)} \geq t\Big) \\
& = \Pr\Big(\big|D^{-1} \sum_{d = 1}^D (X^{(m,d)} - \E_{\tilde \Z^{(m,d)} \mid \Z^{(n)}} X^{(m,d)})\big| \geq t\Big)\\
 & \leq 2 \exp(-s D t) 
\E_{\Z^{(n)}} \prod_{d=1}^D \E_{\tilde \Z^{(m,d)} \mid \Z^{(n)}} \big(
\exp\big(s(X^{(m,d)} - \E_{\tilde \Z^{(m,d)} \mid \Z^{(n)}} X^{(m,d)})\big) \big)
\\
 & \leq 2 \exp(-s D t) \exp(D s^2 / 8) \leq 2 \exp(-2 D t^2),
\end{align*}
}
where $\Pr$ denotes the probability, taking into account all sources of randomness.

For any $\delta \in (0, 1)$, by choosing $t = \sqrt{\frac{\log \frac{2}{\delta}}{2D}}$, 
we obtain that $\abs{F^D_{\tilde T}(x) - F_{\tilde T}(x)} \leq \sqrt{\frac{\log \frac{2}{\delta}}{2D}}$, with probability at least $1 - \delta$.
On the other hand, $\abs{F_{\tilde T}(x) - F_T(x)} \leq \text{TV}(\tilde{\Z}^{(m)}, \Z^{(m)})$ given $\mb Z^{(n)}$ for any $x \in \mc R$, where $\mb Z^{(m)}$ is a sample of size $m$ from $F$.
Using the union bound to combine these results, we obtain that
\begin{equation}
\label{bounds}
    \sup_{\bm x} \abs{F_{\tilde T}(\bm x) - F_T(\bm x)} \leq \sqrt{\frac{\log \frac{2}{\delta}}{2D}} +
\text{TV}(\tilde{\Z}^{(m)}, \Z^{(m)}),
\end{equation}
with probability at least $1 - \delta$. Note that $\text{TV}(\tilde{\Z}^{(m)}, \Z)=
m \cdot \text{TV}(\tilde F, F)$. 

Consequently, as $m \cdot \text{TV}(\tilde F, F) \rightarrow 0$ in probability and $D \rightarrow \infty$,
$\sup_{\bm x} \abs{\hF_T(\bm x) - F_T(\bm x)} \rightarrow 0$ in probability.
Using Syn-Test, the control
of the empirical Type-I error at the $\alpha$ level can be obtained as if we were using the true null
distribution.

Concerning the power gain, note that $\tilde \phi_{m, \alpha} - \phi_{n, \alpha} = \tilde \phi_{m, \alpha} - \phi_{m, \alpha} + \Delta$. By definition and assumption, $|\tilde \phi_{m, \alpha} - \phi_{m, \alpha} | \leq \text{TV}(\tilde{\Z}^{(m)}, \Z^{(m)})
=m \text{TV}(\tilde{F}, F) < \Delta$, indicating that $\tilde \phi_{m, \alpha} - \phi_{n, \alpha}  > 0$.
This completes the proof.
\end{proof}

\section{Additional Experiment} \label{appendix: syn-slm}
In this section, we conduct a simulation study on traditional tabular data to demonstrate the efficacy of the Syn-Slm method, as discussed in Section \ref{dist}. Utilizing the model in equation \eqref{eq: simulation-prediction} as the basis for our data generation, we create training data of $1,000$ instances and test data of $500$ instances. For traditional analysis, we train CatBoost regression using the training dataset. For the Syn-Slm approach, a pre-trained TDM, initially pre-trained with pre-training data of size $2,000$, is fine-tuned on the training set. This fine-tuned model is then used for conditional generation \citep{zhang2023mixed} to approximate $\E(\mb Y|\mb X)$ via a Monte Carlo method. Specifically, for any given input $\mb x$, we generate a series of $(\tilde{\mb Y}^{(d)})_{d=1}^D$ from the conditional distribution $\tilde{\mb Y} | \mb X = \mb x$ as learned by the TDM, and compute the estimate as $D^{-1} \sum_{d = 1}^D \tilde{\mb Y}^{(d)}$. A notable advantage of Syn-Slm is its flexibility in handling various supervised learning tasks simultaneously without needing to retrain for different loss functions. For instance, one may use $\tilde{\mb Y} | \mb X = \mb x$  to estimate prediction intervals \citep{liu2023perturbation}, quantiles, or other statistical measures with relative ease. The performance of both methods is assessed on the test dataset using RMSE as the metric. As indicated in Table \ref{tab: simulate-syn-slm}, the Syn-Slm approach consistently outperforms the traditional CatBoost method across various noise levels $\sigma$, showcasing its superior predictive capability.

\begin{table}[H]
\centering
\caption{RMSE of CatBoost and Syn-Slm across different noise magnitude $\sigma \in \{0.1,~0.2,~ 0.5,~1.0\}$. The number inside the parenthesis represents standard error from repeated conditional generations.}
\label{tab: simulate-syn-slm}
\begin{tabular}{@{}lcccc@{}}
\toprule
         & $\sigma = 0.1$           & $\sigma = 0.2$ & $\sigma = 0.5$           & $\sigma = 1.0$           \\ \midrule
CatBoost & 0.1227                   & 0.2245                              & 0.5271                   & 1.0437                   \\
Syn-Slm  & \textbf{0.1202} (0.0005) & \textbf{0.2171} (0.0008)            & \textbf{0.5148} (0.0020) & \textbf{1.0161} (0.0038) \\ \bottomrule
\end{tabular}
\end{table}
\end{appendix}


\begin{funding}
This work was supported in part by NSF grant DMS-1952539 and NIH grants
R01AG069895, R01AG065636, R01AG074858, U01AG073079 (corresponding author: Xiaotong Shen, \href{mailto:xshen@umn.edu}{xshen@umn.edu}).
\end{funding}

\begin{supplement}
\label{sec:supplement}
\noindent The code to reproduce the results presented in this paper is available at 
 \href{https://github.com/yifei-liu-stat/syn}{https://github.com/yifei-liu-stat/syn}.

\stitle{Datasets}
\sdescription{
Information concerning datasets in Section \ref{numerical}: Case studies.
\begin{table}[H]
\centering
\caption{Summaries of eight Benchmark Datasets \citep{kotelnikov2023tabddpm}.}
\label{tab:dataset-meta-data}
\scalebox{0.74}{
\begin{tabular}{@{}lcccccccc@{}}
\toprule
Dataset                      & FB comments & Adult & House & California & Churn2 & Gesture & Aabalone & Insurance \\ \midrule
Source          & \citeyear{facebook} & \citeyear{adult} & \href{https://www.openml.org/search?type=data\&sort=runs\&id=574\&status=active}{OpenML}       & \citeyear{california} & \href{https://www.kaggle.com/datasets/shrutimechlearn/churn-modelling}{Kaggle}      & \citeyear{gesture} & \href{https://www.openml.org/search?type=data\&sort=runs\&id=183\&status=active}{OpenML}      & \href{https://www.kaggle.com/datasets/mirichoi0218/insurance}{Kaggle}     \\
Sample size           & 197080      & 48842 & 22784 & 20640      & 10000  & 9873    & 4177    & 1338      \\
Pre-training size   & 157638 (80\%)   & 26048 (53\%) & 14581 (64\%) & 13209 (64\%)      & 6400 (64\%) & 6318 (64\%)    & 2672 (64\%) & 856 (64\%) \\
Raw size       & 19720 (10\%)    & 16281 (34\%) & 4557 (20\%)  & 4128 (20\%)       & 2000 (20\%) & 1975 (20\%)    & 836 (20\%)  & 268 (20\%) \\
Test size & 19722 (10\%)    & 6513 (13\%)  & 3646 (16\%)  & 3303 (16\%)       & 1600 (16\%) & 1580 (16\%)    & 669 (16\%)  & 214 (16\%) \\
\# Numerical features & 36          & 6     & 16    & 8          & 7      & 32      & 7       & 3         \\
\# Nominal features   & 15          & 8     & 0     & 0          & 4      & 0       & 1       & 3         \\ \bottomrule
\end{tabular}
}
\end{table}
}
\end{supplement}

\begin{supplement}
\stitle{Training Configurations for Sentiment Analysis}
\sdescription{GPT-3.5 is fine-tuned using the text-embedding-Ada-002 configuration, following OpenAI's recommended practices\footnote{OpenAI GPT fine-tuning: \href{https://platform.openai.com/docs/guides/fine-tuning}{https://platform.openai.com/docs/guides/fine-tuning}}. DistilBERT is fine-tuned with the distilbert-base-uncased model, available from HuggingFace\footnote{HuggingFace: \href{https://huggingface.co/distilbert-base-uncased}{https://huggingface.co/distilbert-base-uncased}}, adopting a batch size of 16, a duration of 10 epochs, and employing the Adam optimizer with default decay parameters and a learning rate of $1\times 10^{-5}$. The LSTM setup is adapted from a Kaggle notebook\footnote{Kaggle: \href{https://www.kaggle.com/code/pawan2905/imbd-sentiment-analysis-using-pytorch-lstm}{https://www.kaggle.com/code/pawan2905/imbd-sentiment-analysis-using-pytorch-lstm}}, utilizing the same model architecture and hyperparameters for our custom-split datasets.}
\end{supplement}


\begin{supplement}
\stitle{Training Configurations for Syn-Boost and Syn-Test}
\sdescription{In accordance with the guidelines of \cite{kotelnikov2023tabddpm}, we process data, and design model architecture and configurations to train our TDMs. Training is conducted on a single TITAN RTX GPU.\\ 
\newline
Prior to input into the diffusion model, tabular data undergoes a transformation process. Continuous variables are normalized through quantile transformation, while categorical variables receive one-hot encoding. These processed variables are then concatenated to form the diffusion model's initial input vector. \\
\newline
For training the models in our examples, we employ a FNN as the backbone of the diffusion model, along with a cosine scheduler \citep{nichol2021improved} and the AdamW optimizer \citep{loshchilov2017decoupled} for optimization. In real-world scenarios involving eight benchmark datasets in Sections \ref{prediction-real} and \ref{sec: inference-real}, we develop pre-trained models using the configurations in Table \ref{tab: training-details}. Model fine-tuning followed the same architectural design but typically involved lighter optimization — a reduced learning rate and fewer training iterations. For the simulation examples in Sections \ref{prediction-real} and \ref{sec: inference-real}, we derive pre-trained models using the same pre-training configuration as that on the California dataset but with $50,000$ training iterations. Fine-tuning in these cases was carried out with $1,000$ iterations and a learning rate of $3 \times 10^{-6}$.
\begin{table}[H]
\centering
\caption{Model design and training details on eight real-world benchmark datasets. Adult-Male uses the same pre-training configuration as Adult does.}
\label{tab: training-details}
\resizebox{\columnwidth}{!}{%
\begin{tabular}{@{}lccccc@{}}
\toprule
Dataset     & FNN dimensions              & Diffusion timestamps & Iterations & Learning rate        & Batch size \\ \midrule
FB comments & (512, 1024)                        & 1000                 & 30000      & $6.3 \times 10^{-4}$ & 4096       \\
Adult       & (256, 1024, 1024, 1024, 1024, 256) & 100                  & 30000      & $2.0 \times 10^{-3}$ & 4096       \\
House       & (128, 512, 512, 512, 512, 256)     & 1000                 & 30000      & $1.4 \times 10^{-3}$ & 4096       \\
California  & (512, 256, 256, 256, 256, 128)     & 1000                 & 30000      & $1.3 \times 10^{-4}$ & 4096       \\
Churn2      & (512, 1024, 1024, 1024, 1024, 512) & 100                  & 30000      & $8.1 \times 10^{-4}$ & 4096       \\
Gesture     & (128, 512, 512, 1024)              & 1000                 & 30000      & $2.8 \times 10^{-3}$ & 4096       \\
Abalone     & (512, 128)                         & 100                  & 30000      & $3.4 \times 10^{-4}$ & 4096       \\
Insurance   & (256, 512, 512, 512, 512, 256)     & 100                  & 30000      & $1.1 \times 10^{-3}$ & 4096       \\ \bottomrule
\end{tabular}%
}
\end{table}
}
\end{supplement}


\bibliographystyle{imsart-nameyear} 
\bibliography{preref-1,diffpriv-1}       

\begin{thebibliography}{60}

\bibitem[\protect\citeauthoryear{Abadi et~al.}{2016}]{abadi2016deep}
\begin{binproceedings}[author]
\bauthor{\bsnm{Abadi},~\bfnm{Martin}\binits{M.}},
  \bauthor{\bsnm{Chu},~\bfnm{Andy}\binits{A.}},
  \bauthor{\bsnm{Goodfellow},~\bfnm{Ian}\binits{I.}},
  \bauthor{\bsnm{McMahan},~\bfnm{H~Brendan}\binits{H.~B.}},
  \bauthor{\bsnm{Mironov},~\bfnm{Ilya}\binits{I.}},
  \bauthor{\bsnm{Talwar},~\bfnm{Kunal}\binits{K.}} \AND
  \bauthor{\bsnm{Zhang},~\bfnm{Li}\binits{L.}}
(\byear{2016}).
\btitle{Deep learning with differential privacy}.
In \bbooktitle{Proceedings of the 2016 ACM SIGSAC conference on computer and
  communications security}
\bpages{308--318}.
\end{binproceedings}
\endbibitem

\bibitem[\protect\citeauthoryear{Akbar, Wang and
  Eklund}{2023}]{akbar2023beware}
\begin{barticle}[author]
\bauthor{\bsnm{Akbar},~\bfnm{Muhammad~Usman}\binits{M.~U.}},
  \bauthor{\bsnm{Wang},~\bfnm{Wuhao}\binits{W.}} \AND
  \bauthor{\bsnm{Eklund},~\bfnm{Anders}\binits{A.}}
(\byear{2023}).
\btitle{Beware of Diffusion Models for Synthesizing Medical Images-a Comparison
  with Gans in Terms of Memorizing Brain MRI and Chest X-Ray Images}.
\bjournal{Available at SSRN 4611613}.
\end{barticle}
\endbibitem

\bibitem[\protect\citeauthoryear{Breiman}{1997}]{breiman1997arcing}
\begin{btechreport}[author]
\bauthor{\bsnm{Breiman},~\bfnm{Leo}\binits{L.}}
(\byear{1997}).
\btitle{Arcing the edge}
\btype{Technical Report},
\bpublisher{Citeseer}.
\end{btechreport}
\endbibitem

\bibitem[\protect\citeauthoryear{Brown et~al.}{2020}]{brown2020language}
\begin{barticle}[author]
\bauthor{\bsnm{Brown},~\bfnm{Tom}\binits{T.}},
  \bauthor{\bsnm{Mann},~\bfnm{Benjamin}\binits{B.}},
  \bauthor{\bsnm{Ryder},~\bfnm{Nick}\binits{N.}},
  \bauthor{\bsnm{Subbiah},~\bfnm{Melanie}\binits{M.}},
  \bauthor{\bsnm{Kaplan},~\bfnm{Jared~D}\binits{J.~D.}},
  \bauthor{\bsnm{Dhariwal},~\bfnm{Prafulla}\binits{P.}},
  \bauthor{\bsnm{Neelakantan},~\bfnm{Arvind}\binits{A.}},
  \bauthor{\bsnm{Shyam},~\bfnm{Pranav}\binits{P.}},
  \bauthor{\bsnm{Sastry},~\bfnm{Girish}\binits{G.}},
  \bauthor{\bsnm{Askell},~\bfnm{Amanda}\binits{A.}} \betal{et~al.}
(\byear{2020}).
\btitle{Language models are few-shot learners}.
\bjournal{Advances in neural information processing systems}
\bvolume{33}
\bpages{1877--1901}.
\end{barticle}
\endbibitem

\bibitem[\protect\citeauthoryear{Bubeck et~al.}{2023}]{bubeck2023sparks}
\begin{barticle}[author]
\bauthor{\bsnm{Bubeck},~\bfnm{S{\'e}bastien}\binits{S.}},
  \bauthor{\bsnm{Chandrasekaran},~\bfnm{Varun}\binits{V.}},
  \bauthor{\bsnm{Eldan},~\bfnm{Ronen}\binits{R.}},
  \bauthor{\bsnm{Gehrke},~\bfnm{Johannes}\binits{J.}},
  \bauthor{\bsnm{Horvitz},~\bfnm{Eric}\binits{E.}},
  \bauthor{\bsnm{Kamar},~\bfnm{Ece}\binits{E.}},
  \bauthor{\bsnm{Lee},~\bfnm{Peter}\binits{P.}},
  \bauthor{\bsnm{Lee},~\bfnm{Yin~Tat}\binits{Y.~T.}},
  \bauthor{\bsnm{Li},~\bfnm{Yuanzhi}\binits{Y.}},
  \bauthor{\bsnm{Lundberg},~\bfnm{Scott}\binits{S.}} \betal{et~al.}
(\byear{2023}).
\btitle{Sparks of artificial general intelligence: Early experiments with
  gpt-4}.
\bjournal{arXiv preprint arXiv:2303.12712}.
\end{barticle}
\endbibitem

\bibitem[\protect\citeauthoryear{Burden and Faires}{2001}]{burden2001numerical}
\begin{bmisc}[author]
\bauthor{\bsnm{Burden},~\bfnm{RL}\binits{R.}} \AND
  \bauthor{\bsnm{Faires},~\bfnm{JD}\binits{J.}}
(\byear{2001}).
\btitle{Numerical analysis 7th ed., brooks/cole, thomson learning}.
\end{bmisc}
\endbibitem

\bibitem[\protect\citeauthoryear{Carroll}{2006}]{carroll2006movement}
\begin{barticle}[author]
\bauthor{\bsnm{Carroll},~\bfnm{Michael~W}\binits{M.~W.}}
(\byear{2006}).
\btitle{The movement for open access law}.
\bjournal{Law Library Journal}
\bvolume{92}
\bpages{315}.
\end{barticle}
\endbibitem

\bibitem[\protect\citeauthoryear{Chen et~al.}{2023}]{chen2023robust}
\begin{barticle}[author]
\bauthor{\bsnm{Chen},~\bfnm{Xuanting}\binits{X.}},
  \bauthor{\bsnm{Ye},~\bfnm{Junjie}\binits{J.}},
  \bauthor{\bsnm{Zu},~\bfnm{Can}\binits{C.}},
  \bauthor{\bsnm{Xu},~\bfnm{Nuo}\binits{N.}},
  \bauthor{\bsnm{Zheng},~\bfnm{Rui}\binits{R.}},
  \bauthor{\bsnm{Peng},~\bfnm{Minlong}\binits{M.}},
  \bauthor{\bsnm{Zhou},~\bfnm{Jie}\binits{J.}},
  \bauthor{\bsnm{Gui},~\bfnm{Tao}\binits{T.}},
  \bauthor{\bsnm{Zhang},~\bfnm{Qi}\binits{Q.}} \AND
  \bauthor{\bsnm{Huang},~\bfnm{Xuanjing}\binits{X.}}
(\byear{2023}).
\btitle{How Robust is GPT-3.5 to Predecessors? A Comprehensive Study on
  Language Understanding Tasks}.
\bjournal{arXiv preprint arXiv:2303.00293}.
\end{barticle}
\endbibitem

\bibitem[\protect\citeauthoryear{Dai, Shen and Pan}{2024}]{dai2024significance}
\begin{barticle}[author]
\bauthor{\bsnm{Dai},~\bfnm{Ben}\binits{B.}},
  \bauthor{\bsnm{Shen},~\bfnm{Xiaotong}\binits{X.}} \AND
  \bauthor{\bsnm{Pan},~\bfnm{Wei}\binits{W.}}
(\byear{2024}).
\btitle{Significance tests of feature relevance for a black-box learner}.
\bjournal{IEEE Transactions on Neural Networks and Learning Systems}
\bvolume{35}
\bpages{1898-1911}.
\end{barticle}
\endbibitem

\bibitem[\protect\citeauthoryear{Dinh, Krueger and Bengio}{2014}]{dinh2014nice}
\begin{barticle}[author]
\bauthor{\bsnm{Dinh},~\bfnm{Laurent}\binits{L.}},
  \bauthor{\bsnm{Krueger},~\bfnm{David}\binits{D.}} \AND
  \bauthor{\bsnm{Bengio},~\bfnm{Yoshua}\binits{Y.}}
(\byear{2014}).
\btitle{Nice: Non-linear independent components estimation}.
\bjournal{arXiv preprint arXiv:1410.8516}.
\end{barticle}
\endbibitem

\bibitem[\protect\citeauthoryear{Dinh, Sohl-Dickstein and
  Bengio}{2016}]{dinh2016density}
\begin{barticle}[author]
\bauthor{\bsnm{Dinh},~\bfnm{Laurent}\binits{L.}},
  \bauthor{\bsnm{Sohl-Dickstein},~\bfnm{Jascha}\binits{J.}} \AND
  \bauthor{\bsnm{Bengio},~\bfnm{Samy}\binits{S.}}
(\byear{2016}).
\btitle{Density estimation using real nvp}.
\bjournal{arXiv preprint arXiv:1605.08803}.
\end{barticle}
\endbibitem

\bibitem[\protect\citeauthoryear{Dorogush, Ershov and
  Gulin}{2018}]{dorogush2018catboost}
\begin{barticle}[author]
\bauthor{\bsnm{Dorogush},~\bfnm{Anna~V}\binits{A.~V.}},
  \bauthor{\bsnm{Ershov},~\bfnm{Vadim}\binits{V.}} \AND
  \bauthor{\bsnm{Gulin},~\bfnm{Andrey}\binits{A.}}
(\byear{2018}).
\btitle{CatBoost: unbiased boosting with categorical features}.
\bjournal{arXiv preprint arXiv:1810.11363}.
\end{barticle}
\endbibitem

\bibitem[\protect\citeauthoryear{Dwork et~al.}{2006}]{dwork2006our}
\begin{binproceedings}[author]
\bauthor{\bsnm{Dwork},~\bfnm{Cynthia}\binits{C.}},
  \bauthor{\bsnm{Kenthapadi},~\bfnm{Krishnaram}\binits{K.}},
  \bauthor{\bsnm{McSherry},~\bfnm{Frank}\binits{F.}},
  \bauthor{\bsnm{Mironov},~\bfnm{Ilya}\binits{I.}} \AND
  \bauthor{\bsnm{Naor},~\bfnm{Moni}\binits{M.}}
(\byear{2006}).
\btitle{Our data, ourselves: Privacy via distributed noise generation}.
In \bbooktitle{Annual International Conference on the Theory and Applications
  of Cryptographic Techniques}
\bpages{486--503}.
\end{binproceedings}
\endbibitem

\bibitem[\protect\citeauthoryear{Eastwood}{2023}]{mitreport}
\begin{barticle}[author]
\bauthor{\bsnm{Eastwood},~\bfnm{Brian}\binits{B.}}
(\byear{2023}).
\btitle{What is synthetic data — and how can it help you competitively?}
\bjournal{MIT Sloan School}.
\end{barticle}
\endbibitem

\bibitem[\protect\citeauthoryear{Efron}{1992}]{efron1992bootstrap}
\begin{bincollection}[author]
\bauthor{\bsnm{Efron},~\bfnm{Bradley}\binits{B.}}
(\byear{1992}).
\btitle{Bootstrap methods: another look at the jackknife}.
In \bbooktitle{Breakthroughs in statistics}
\bpages{569--593}.
\bpublisher{Springer}.
\end{bincollection}
\endbibitem

\bibitem[\protect\citeauthoryear{Frid-Adar et~al.}{2018}]{frid2018gan}
\begin{barticle}[author]
\bauthor{\bsnm{Frid-Adar},~\bfnm{Maayan}\binits{M.}},
  \bauthor{\bsnm{Diamant},~\bfnm{Idit}\binits{I.}},
  \bauthor{\bsnm{Klang},~\bfnm{Eyal}\binits{E.}},
  \bauthor{\bsnm{Amitai},~\bfnm{Michal}\binits{M.}},
  \bauthor{\bsnm{Goldberger},~\bfnm{Jacob}\binits{J.}} \AND
  \bauthor{\bsnm{Greenspan},~\bfnm{Hayit}\binits{H.}}
(\byear{2018}).
\btitle{GAN-based synthetic medical image augmentation for increased CNN
  performance in liver lesion classification}.
\bjournal{Neurocomputing}
\bvolume{321}
\bpages{321--331}.
\end{barticle}
\endbibitem

\bibitem[\protect\citeauthoryear{Friedman}{2002}]{friedman2002stochastic}
\begin{barticle}[author]
\bauthor{\bsnm{Friedman},~\bfnm{Jerome~H}\binits{J.~H.}}
(\byear{2002}).
\btitle{Stochastic gradient boosting}.
\bjournal{Computational statistics \& data analysis}
\bvolume{38}
\bpages{367--378}.
\end{barticle}
\endbibitem

\bibitem[\protect\citeauthoryear{Gao et~al.}{2023}]{gao2023synthetic}
\begin{barticle}[author]
\bauthor{\bsnm{Gao},~\bfnm{Cong}\binits{C.}},
  \bauthor{\bsnm{Killeen},~\bfnm{Benjamin~D}\binits{B.~D.}},
  \bauthor{\bsnm{Hu},~\bfnm{Yicheng}\binits{Y.}},
  \bauthor{\bsnm{Grupp},~\bfnm{Robert~B}\binits{R.~B.}},
  \bauthor{\bsnm{Taylor},~\bfnm{Russell~H}\binits{R.~H.}},
  \bauthor{\bsnm{Armand},~\bfnm{Mehran}\binits{M.}} \AND
  \bauthor{\bsnm{Unberath},~\bfnm{Mathias}\binits{M.}}
(\byear{2023}).
\btitle{Synthetic data accelerates the development of generalizable
  learning-based algorithms for X-ray image analysis}.
\bjournal{Nature Machine Intelligence}
\bvolume{5}
\bpages{294--308}.
\end{barticle}
\endbibitem

\bibitem[\protect\citeauthoryear{Gartner}{2022}]{gartner}
\begin{barticle}[author]
\bauthor{\bsnm{Gartner}}
(\byear{2022}).
\btitle{Is Synthetic Data the Future of AI?}
\bjournal{Gartner Newsroom Q\&A}.
\end{barticle}
\endbibitem

\bibitem[\protect\citeauthoryear{Ghalebikesabi
  et~al.}{2023}]{ghalebikesabi2023differentiall}
\begin{barticle}[author]
\bauthor{\bsnm{Ghalebikesabi},~\bfnm{Sahra}\binits{S.}},
  \bauthor{\bsnm{Berrada},~\bfnm{Leonard}\binits{L.}},
  \bauthor{\bsnm{Gowal},~\bfnm{Sven}\binits{S.}},
  \bauthor{\bsnm{Ktena},~\bfnm{Ira}\binits{I.}},
  \bauthor{\bsnm{Stanforth},~\bfnm{Robert}\binits{R.}},
  \bauthor{\bsnm{Hayes},~\bfnm{Jamie}\binits{J.}},
  \bauthor{\bsnm{De},~\bfnm{Soham}\binits{S.}},
  \bauthor{\bsnm{Smith},~\bfnm{Samuel~L}\binits{S.~L.}},
  \bauthor{\bsnm{Wiles},~\bfnm{Olivia}\binits{O.}} \AND
  \bauthor{\bsnm{Balle},~\bfnm{Borja}\binits{B.}}
(\byear{2023}).
\btitle{Differentially private diffusion models generate useful synthetic
  images}.
\bjournal{arXiv preprint arXiv:2302.13861}.
\end{barticle}
\endbibitem

\bibitem[\protect\citeauthoryear{Hastie et~al.}{2009}]{hastie2009elements}
\begin{bbook}[author]
\bauthor{\bsnm{Hastie},~\bfnm{Trevor}\binits{T.}},
  \bauthor{\bsnm{Tibshirani},~\bfnm{Robert}\binits{R.}},
  \bauthor{\bsnm{Friedman},~\bfnm{Jerome~H}\binits{J.~H.}} \AND
  \bauthor{\bsnm{Friedman},~\bfnm{Jerome~H}\binits{J.~H.}}
(\byear{2009}).
\btitle{The elements of statistical learning: data mining, inference, and
  prediction}
\bvolume{2}.
\bpublisher{Springer}.
\end{bbook}
\endbibitem

\bibitem[\protect\citeauthoryear{Ho, Jain and Abbeel}{2020}]{ho2020denoising}
\begin{barticle}[author]
\bauthor{\bsnm{Ho},~\bfnm{Jonathan}\binits{J.}},
  \bauthor{\bsnm{Jain},~\bfnm{Ajay}\binits{A.}} \AND
  \bauthor{\bsnm{Abbeel},~\bfnm{Pieter}\binits{P.}}
(\byear{2020}).
\btitle{Denoising diffusion probabilistic models}.
\bjournal{Advances in Neural Information Processing Systems}
\bvolume{33}
\bpages{6840--6851}.
\end{barticle}
\endbibitem

\bibitem[\protect\citeauthoryear{Hommel}{1983}]{hommel1983tests}
\begin{barticle}[author]
\bauthor{\bsnm{Hommel},~\bfnm{Gerhard}\binits{G.}}
(\byear{1983}).
\btitle{Tests of the overall hypothesis for arbitrary dependence structures}.
\bjournal{Biometrical Journal}
\bvolume{25}
\bpages{423--430}.
\end{barticle}
\endbibitem

\bibitem[\protect\citeauthoryear{Jordon et~al.}{2022}]{jordon2022synthetic}
\begin{barticle}[author]
\bauthor{\bsnm{Jordon},~\bfnm{James}\binits{J.}},
  \bauthor{\bsnm{Szpruch},~\bfnm{Lukasz}\binits{L.}},
  \bauthor{\bsnm{Houssiau},~\bfnm{Florimond}\binits{F.}},
  \bauthor{\bsnm{Bottarelli},~\bfnm{Mirko}\binits{M.}},
  \bauthor{\bsnm{Cherubin},~\bfnm{Giovanni}\binits{G.}},
  \bauthor{\bsnm{Maple},~\bfnm{Carsten}\binits{C.}},
  \bauthor{\bsnm{Cohen},~\bfnm{Samuel~N}\binits{S.~N.}} \AND
  \bauthor{\bsnm{Weller},~\bfnm{Adrian}\binits{A.}}
(\byear{2022}).
\btitle{Synthetic Data--what, why and how?}
\bjournal{arXiv preprint arXiv:2205.03257}.
\end{barticle}
\endbibitem

\bibitem[\protect\citeauthoryear{Kiefer}{1953}]{kiefer1953sequential}
\begin{barticle}[author]
\bauthor{\bsnm{Kiefer},~\bfnm{Jack}\binits{J.}}
(\byear{1953}).
\btitle{Sequential minimax search for a maximum}.
\bjournal{Proceedings of the American mathematical society}
\bvolume{4}
\bpages{502--506}.
\end{barticle}
\endbibitem

\bibitem[\protect\citeauthoryear{Kim, Lee and Park}{2022}]{kim2022stasy}
\begin{barticle}[author]
\bauthor{\bsnm{Kim},~\bfnm{Jayoung}\binits{J.}},
  \bauthor{\bsnm{Lee},~\bfnm{Chaejeong}\binits{C.}} \AND
  \bauthor{\bsnm{Park},~\bfnm{Noseong}\binits{N.}}
(\byear{2022}).
\btitle{Stasy: Score-based tabular data synthesis}.
\bjournal{arXiv preprint arXiv:2210.04018}.
\end{barticle}
\endbibitem

\bibitem[\protect\citeauthoryear{Kingma and Dhariwal}{2018}]{kingma2018glow}
\begin{barticle}[author]
\bauthor{\bsnm{Kingma},~\bfnm{Durk~P}\binits{D.~P.}} \AND
  \bauthor{\bsnm{Dhariwal},~\bfnm{Prafulla}\binits{P.}}
(\byear{2018}).
\btitle{Glow: Generative flow with invertible 1x1 convolutions}.
\bjournal{Advances in neural information processing systems}
\bvolume{31}.
\end{barticle}
\endbibitem

\bibitem[\protect\citeauthoryear{Kohavi et~al.}{1996}]{adult}
\begin{binproceedings}[author]
\bauthor{\bsnm{Kohavi},~\bfnm{Ron}\binits{R.}} \betal{et~al.}
(\byear{1996}).
\btitle{Scaling up the accuracy of naive-bayes classifiers: A decision-tree
  hybrid.}
In \bbooktitle{Kdd}
\bvolume{96}
\bpages{202--207}.
\end{binproceedings}
\endbibitem

\bibitem[\protect\citeauthoryear{Kotelnikov
  et~al.}{2023}]{kotelnikov2023tabddpm}
\begin{binproceedings}[author]
\bauthor{\bsnm{Kotelnikov},~\bfnm{Akim}\binits{A.}},
  \bauthor{\bsnm{Baranchuk},~\bfnm{Dmitry}\binits{D.}},
  \bauthor{\bsnm{Rubachev},~\bfnm{Ivan}\binits{I.}} \AND
  \bauthor{\bsnm{Babenko},~\bfnm{Artem}\binits{A.}}
(\byear{2023}).
\btitle{Tabddpm: Modelling tabular data with diffusion models}.
In \bbooktitle{International Conference on Machine Learning}
\bpages{17564--17579}.
\bpublisher{PMLR}.
\end{binproceedings}
\endbibitem

\bibitem[\protect\citeauthoryear{Labrinidis and
  Jagadish}{2012}]{labrinidis2012challenges}
\begin{barticle}[author]
\bauthor{\bsnm{Labrinidis},~\bfnm{Alexandros}\binits{A.}} \AND
  \bauthor{\bsnm{Jagadish},~\bfnm{HV}\binits{H.}}
(\byear{2012}).
\btitle{Challenges and opportunities with big data}.
\bjournal{Proceedings of the VLDB Endowment}
\bvolume{5}
\bpages{2032--2033}.
\end{barticle}
\endbibitem

\bibitem[\protect\citeauthoryear{Lee, Kim and Park}{2023}]{lee2023codi}
\begin{barticle}[author]
\bauthor{\bsnm{Lee},~\bfnm{Chaejeong}\binits{C.}},
  \bauthor{\bsnm{Kim},~\bfnm{Jayoung}\binits{J.}} \AND
  \bauthor{\bsnm{Park},~\bfnm{Noseong}\binits{N.}}
(\byear{2023}).
\btitle{CoDi: Co-evolving Contrastive Diffusion Models for Mixed-type Tabular
  Synthesis}.
\bjournal{arXiv preprint arXiv:2304.12654}.
\end{barticle}
\endbibitem

\bibitem[\protect\citeauthoryear{Lin et~al.}{2023}]{lin2023diffusion}
\begin{barticle}[author]
\bauthor{\bsnm{Lin},~\bfnm{Lequan}\binits{L.}},
  \bauthor{\bsnm{Li},~\bfnm{Zhengkun}\binits{Z.}},
  \bauthor{\bsnm{Li},~\bfnm{Ruikun}\binits{R.}},
  \bauthor{\bsnm{Li},~\bfnm{Xuliang}\binits{X.}} \AND
  \bauthor{\bsnm{Gao},~\bfnm{Junbin}\binits{J.}}
(\byear{2023}).
\btitle{Diffusion models for time series applications: A survey}.
\bjournal{arXiv preprint arXiv:2305.00624}.
\end{barticle}
\endbibitem

\bibitem[\protect\citeauthoryear{Liu, Shen and
  Shen}{2023}]{liu2023perturbation}
\begin{barticle}[author]
\bauthor{\bsnm{Liu},~\bfnm{Yifei}\binits{Y.}},
  \bauthor{\bsnm{Shen},~\bfnm{Rex}\binits{R.}} \AND
  \bauthor{\bsnm{Shen},~\bfnm{Xiaotong}\binits{X.}}
(\byear{2023}).
\btitle{Perturbation-Assisted Sample Synthesis: A Novel Approach for
  Uncertainty Quantification}.
\bjournal{Revised for {\it IEEE Transactions on Pattern Analysis and Machine
  Intelligence}i. arXiv preprint arXiv:2305.18671}.
\end{barticle}
\endbibitem

\bibitem[\protect\citeauthoryear{Liu and Xie}{2020}]{liu2020cauchy}
\begin{barticle}[author]
\bauthor{\bsnm{Liu},~\bfnm{Yaowu}\binits{Y.}} \AND
  \bauthor{\bsnm{Xie},~\bfnm{Jun}\binits{J.}}
(\byear{2020}).
\btitle{Cauchy combination test: a powerful test with analytic p-value
  calculation under arbitrary dependency structures}.
\bjournal{Journal of the American Statistical Association}
\bvolume{115}
\bpages{393--402}.
\end{barticle}
\endbibitem

\bibitem[\protect\citeauthoryear{Liu et~al.}{2020}]{liu2020roundtrip}
\begin{barticle}[author]
\bauthor{\bsnm{Liu},~\bfnm{Qiao}\binits{Q.}},
  \bauthor{\bsnm{Xu},~\bfnm{Jiaze}\binits{J.}},
  \bauthor{\bsnm{Jiang},~\bfnm{Rui}\binits{R.}} \AND
  \bauthor{\bsnm{Wong},~\bfnm{Wing~Hung}\binits{W.~H.}}
(\byear{2020}).
\btitle{Roundtrip: A Deep Generative Neural Density Estimator}.
\bjournal{arXiv preprint arXiv:2004.09017}.
\end{barticle}
\endbibitem

\bibitem[\protect\citeauthoryear{Loshchilov and
  Hutter}{2017}]{loshchilov2017decoupled}
\begin{barticle}[author]
\bauthor{\bsnm{Loshchilov},~\bfnm{Ilya}\binits{I.}} \AND
  \bauthor{\bsnm{Hutter},~\bfnm{Frank}\binits{F.}}
(\byear{2017}).
\btitle{Decoupled weight decay regularization}.
\bjournal{arXiv preprint arXiv:1711.05101}.
\end{barticle}
\endbibitem

\bibitem[\protect\citeauthoryear{Maas et~al.}{2011}]{maas2011learning}
\begin{binproceedings}[author]
\bauthor{\bsnm{Maas},~\bfnm{Andrew}\binits{A.}},
  \bauthor{\bsnm{Daly},~\bfnm{Raymond~E}\binits{R.~E.}},
  \bauthor{\bsnm{Pham},~\bfnm{Peter~T}\binits{P.~T.}},
  \bauthor{\bsnm{Huang},~\bfnm{Dan}\binits{D.}},
  \bauthor{\bsnm{Ng},~\bfnm{Andrew~Y}\binits{A.~Y.}} \AND
  \bauthor{\bsnm{Potts},~\bfnm{Christopher}\binits{C.}}
(\byear{2011}).
\btitle{Learning word vectors for sentiment analysis}.
In \bbooktitle{Proceedings of the 49th annual meeting of the association for
  computational linguistics: Human language technologies}
\bpages{142--150}.
\end{binproceedings}
\endbibitem

\bibitem[\protect\citeauthoryear{Madeo, Lima and Peres}{2013}]{gesture}
\begin{binproceedings}[author]
\bauthor{\bsnm{Madeo},~\bfnm{Renata~CB}\binits{R.~C.}},
  \bauthor{\bsnm{Lima},~\bfnm{Clodoaldo~AM}\binits{C.~A.}} \AND
  \bauthor{\bsnm{Peres},~\bfnm{Sarajane~M}\binits{S.~M.}}
(\byear{2013}).
\btitle{Gesture unit segmentation using support vector machines: segmenting
  gestures from rest positions}.
In \bbooktitle{Proceedings of the 28th Annual ACM Symposium on Applied
  Computing}
\bpages{46--52}.
\end{binproceedings}
\endbibitem

\bibitem[\protect\citeauthoryear{Nichol and
  Dhariwal}{2021}]{nichol2021improved}
\begin{binproceedings}[author]
\bauthor{\bsnm{Nichol},~\bfnm{Alexander~Quinn}\binits{A.~Q.}} \AND
  \bauthor{\bsnm{Dhariwal},~\bfnm{Prafulla}\binits{P.}}
(\byear{2021}).
\btitle{Improved denoising diffusion probabilistic models}.
In \bbooktitle{International Conference on Machine Learning}
\bpages{8162--8171}.
\bpublisher{PMLR}.
\end{binproceedings}
\endbibitem

\bibitem[\protect\citeauthoryear{Oko, Akiyama and
  Suzuki}{2023}]{oko2023diffusion}
\begin{barticle}[author]
\bauthor{\bsnm{Oko},~\bfnm{Kazusato}\binits{K.}},
  \bauthor{\bsnm{Akiyama},~\bfnm{Shunta}\binits{S.}} \AND
  \bauthor{\bsnm{Suzuki},~\bfnm{Taiji}\binits{T.}}
(\byear{2023}).
\btitle{Diffusion Models are Minimax Optimal Distribution Estimators}.
\bjournal{arXiv preprint arXiv:2303.01861}.
\end{barticle}
\endbibitem

\bibitem[\protect\citeauthoryear{OpenAI}{2023}]{openai2023gpt4}
\begin{bmisc}[author]
\bauthor{\bsnm{OpenAI}}
(\byear{2023}).
\btitle{GPT-4 Technical Report}.
\end{bmisc}
\endbibitem

\bibitem[\protect\citeauthoryear{Pace and Barry}{1997}]{california}
\begin{barticle}[author]
\bauthor{\bsnm{Pace},~\bfnm{R~Kelley}\binits{R.~K.}} \AND
  \bauthor{\bsnm{Barry},~\bfnm{Ronald}\binits{R.}}
(\byear{1997}).
\btitle{Sparse spatial autoregressions}.
\bjournal{Statistics \& Probability Letters}
\bvolume{33}
\bpages{291--297}.
\end{barticle}
\endbibitem

\bibitem[\protect\citeauthoryear{Quionero-Candela
  et~al.}{2009}]{quionero2009dataset}
\begin{bbook}[author]
\beditor{\bsnm{Quionero-Candela},~\bfnm{Joaquin}\binits{J.}},
  \beditor{\bsnm{Sugiyama},~\bfnm{Masashi}\binits{M.}},
  \beditor{\bsnm{Schwaighofer},~\bfnm{Anton}\binits{A.}} \AND
  \beditor{\bsnm{Lawrence},~\bfnm{Neil~D}\binits{N.~D.}}, eds.
(\byear{2009}).
\btitle{Dataset shift in machine learning}.
\bpublisher{The MIT Press}.
\end{bbook}
\endbibitem

\bibitem[\protect\citeauthoryear{Sanh et~al.}{2019}]{sanh2019distilbert}
\begin{barticle}[author]
\bauthor{\bsnm{Sanh},~\bfnm{Victor}\binits{V.}},
  \bauthor{\bsnm{Debut},~\bfnm{Lysandre}\binits{L.}},
  \bauthor{\bsnm{Chaumond},~\bfnm{Julien}\binits{J.}} \AND
  \bauthor{\bsnm{Wolf},~\bfnm{Thomas}\binits{T.}}
(\byear{2019}).
\btitle{DistilBERT, a distilled version of BERT: smaller, faster, cheaper and
  lighter}.
\bjournal{arXiv preprint arXiv:1910.01108}.
\end{barticle}
\endbibitem

\bibitem[\protect\citeauthoryear{Schapire}{1990}]{schapire1990strength}
\begin{barticle}[author]
\bauthor{\bsnm{Schapire},~\bfnm{Robert~E}\binits{R.~E.}}
(\byear{1990}).
\btitle{The strength of weak learnability}.
\bjournal{Machine learning}
\bvolume{5}
\bpages{197--227}.
\end{barticle}
\endbibitem

\bibitem[\protect\citeauthoryear{Shwartz-Ziv and
  Armon}{2022}]{shwartz2022tabular}
\begin{barticle}[author]
\bauthor{\bsnm{Shwartz-Ziv},~\bfnm{Ravid}\binits{R.}} \AND
  \bauthor{\bsnm{Armon},~\bfnm{Amitai}\binits{A.}}
(\byear{2022}).
\btitle{Tabular data: Deep learning is not all you need}.
\bjournal{Information Fusion}
\bvolume{81}
\bpages{84--90}.
\end{barticle}
\endbibitem

\bibitem[\protect\citeauthoryear{Singh, Sandhu and Kumar}{2015}]{facebook}
\begin{binproceedings}[author]
\bauthor{\bsnm{Singh},~\bfnm{Kamaljot}\binits{K.}},
  \bauthor{\bsnm{Sandhu},~\bfnm{Ranjeet~Kaur}\binits{R.~K.}} \AND
  \bauthor{\bsnm{Kumar},~\bfnm{Dinesh}\binits{D.}}
(\byear{2015}).
\btitle{Comment volume prediction using neural networks and decision trees}.
In \bbooktitle{IEEE UKSim-AMSS 17th International Conference on Computer
  Modelling and Simulation, UKSim2015 (UKSim2015)}.
\end{binproceedings}
\endbibitem

\bibitem[\protect\citeauthoryear{Sohl-Dickstein et~al.}{2015}]{sohl2015deep}
\begin{binproceedings}[author]
\bauthor{\bsnm{Sohl-Dickstein},~\bfnm{Jascha}\binits{J.}},
  \bauthor{\bsnm{Weiss},~\bfnm{Eric}\binits{E.}},
  \bauthor{\bsnm{Maheswaranathan},~\bfnm{Niru}\binits{N.}} \AND
  \bauthor{\bsnm{Ganguli},~\bfnm{Surya}\binits{S.}}
(\byear{2015}).
\btitle{Deep unsupervised learning using nonequilibrium thermodynamics}.
In \bbooktitle{International Conference on Machine Learning}
\bpages{2256--2265}.
\bpublisher{PMLR}.
\end{binproceedings}
\endbibitem

\bibitem[\protect\citeauthoryear{Song, Meng and
  Ermon}{2020}]{song2020denoising}
\begin{barticle}[author]
\bauthor{\bsnm{Song},~\bfnm{Jiaming}\binits{J.}},
  \bauthor{\bsnm{Meng},~\bfnm{Chenlin}\binits{C.}} \AND
  \bauthor{\bsnm{Ermon},~\bfnm{Stefano}\binits{S.}}
(\byear{2020}).
\btitle{Denoising Diffusion Implicit Models}.
\bjournal{arXiv:2010.02502}.
\end{barticle}
\endbibitem

\bibitem[\protect\citeauthoryear{Sweeney}{2002}]{sweeney2002kanonymity}
\begin{barticle}[author]
\bauthor{\bsnm{Sweeney},~\bfnm{Latanya}\binits{L.}}
(\byear{2002}).
\btitle{k-anonymity: A model for protecting privacy}.
\bjournal{International Journal of Uncertainty, Fuzziness and Knowledge-Based
  Systems}
\bvolume{10}
\bpages{557--570}.
\end{barticle}
\endbibitem

\bibitem[\protect\citeauthoryear{Toews}{2022}]{toews2022synthetic}
\begin{bmisc}[author]
\bauthor{\bsnm{Toews},~\bfnm{Rob}\binits{R.}}
(\byear{2022}).
\btitle{Synthetic data is about to transform artificial intelligence}.
\end{bmisc}
\endbibitem

\bibitem[\protect\citeauthoryear{Tripuraneni, Jordan and
  Jin}{2020}]{tripuraneni2020theory}
\begin{barticle}[author]
\bauthor{\bsnm{Tripuraneni},~\bfnm{Nilesh}\binits{N.}},
  \bauthor{\bsnm{Jordan},~\bfnm{Michael}\binits{M.}} \AND
  \bauthor{\bsnm{Jin},~\bfnm{Chi}\binits{C.}}
(\byear{2020}).
\btitle{On the theory of transfer learning: The importance of task diversity}.
\bjournal{Advances in neural information processing systems}
\bvolume{33}
\bpages{7852--7862}.
\end{barticle}
\endbibitem

\bibitem[\protect\citeauthoryear{Wasserman, Ramdas and
  Balakrishnan}{2020}]{wasserman2020universal}
\begin{barticle}[author]
\bauthor{\bsnm{Wasserman},~\bfnm{Larry}\binits{L.}},
  \bauthor{\bsnm{Ramdas},~\bfnm{Aaditya}\binits{A.}} \AND
  \bauthor{\bsnm{Balakrishnan},~\bfnm{Sivaraman}\binits{S.}}
(\byear{2020}).
\btitle{Universal inference}.
\bjournal{Proceedings of the National Academy of Sciences}
\bvolume{117}
\bpages{16880--16890}.
\end{barticle}
\endbibitem

\bibitem[\protect\citeauthoryear{Wasserman and
  Roeder}{2009}]{wasserman2009high}
\begin{barticle}[author]
\bauthor{\bsnm{Wasserman},~\bfnm{Larry}\binits{L.}} \AND
  \bauthor{\bsnm{Roeder},~\bfnm{Kathryn}\binits{K.}}
(\byear{2009}).
\btitle{High dimensional variable selection}.
\bjournal{Annals of statistics}
\bvolume{37}
\bpages{2178}.
\end{barticle}
\endbibitem

\bibitem[\protect\citeauthoryear{Wei et~al.}{2021}]{wei2021diffusion}
\begin{barticle}[author]
\bauthor{\bsnm{Wei},~\bfnm{Donglai}\binits{D.}}, \bauthor{} \betal{et~al.}
(\byear{2021}).
\btitle{Diffusion Models Beat GANs on Image Synthesis}.
\bjournal{Preprint arXiv}.
\end{barticle}
\endbibitem

\bibitem[\protect\citeauthoryear{Xu et~al.}{2023}]{xu2023versatile}
\begin{binproceedings}[author]
\bauthor{\bsnm{Xu},~\bfnm{Xingqian}\binits{X.}},
  \bauthor{\bsnm{Wang},~\bfnm{Zhangyang}\binits{Z.}},
  \bauthor{\bsnm{Zhang},~\bfnm{Gong}\binits{G.}},
  \bauthor{\bsnm{Wang},~\bfnm{Kai}\binits{K.}} \AND
  \bauthor{\bsnm{Shi},~\bfnm{Humphrey}\binits{H.}}
(\byear{2023}).
\btitle{Versatile diffusion: Text, images and variations all in one diffusion
  model}.
In \bbooktitle{Proceedings of the IEEE/CVF International Conference on Computer
  Vision}
\bpages{7754--7765}.
\end{binproceedings}
\endbibitem

\bibitem[\protect\citeauthoryear{Yuan et~al.}{2023}]{yuan2023spatio}
\begin{barticle}[author]
\bauthor{\bsnm{Yuan},~\bfnm{Yuan}\binits{Y.}},
  \bauthor{\bsnm{Ding},~\bfnm{Jingtao}\binits{J.}},
  \bauthor{\bsnm{Shao},~\bfnm{Chenyang}\binits{C.}},
  \bauthor{\bsnm{Jin},~\bfnm{Depeng}\binits{D.}} \AND
  \bauthor{\bsnm{Li},~\bfnm{Yong}\binits{Y.}}
(\byear{2023}).
\btitle{Spatio-temporal Diffusion Point Processes}.
\bjournal{arXiv preprint arXiv:2305.12403}.
\end{barticle}
\endbibitem

\bibitem[\protect\citeauthoryear{Zhang et~al.}{2023a}]{zhang2023text}
\begin{barticle}[author]
\bauthor{\bsnm{Zhang},~\bfnm{Chenshuang}\binits{C.}},
  \bauthor{\bsnm{Zhang},~\bfnm{Chaoning}\binits{C.}},
  \bauthor{\bsnm{Zhang},~\bfnm{Mengchun}\binits{M.}} \AND
  \bauthor{\bsnm{Kweon},~\bfnm{In~So}\binits{I.~S.}}
(\byear{2023}a).
\btitle{Text-to-image diffusion model in generative ai: A survey}.
\bjournal{arXiv preprint arXiv:2303.07909}.
\end{barticle}
\endbibitem

\bibitem[\protect\citeauthoryear{Zhang et~al.}{2023b}]{zhang2023mixed}
\begin{barticle}[author]
\bauthor{\bsnm{Zhang},~\bfnm{Hengrui}\binits{H.}},
  \bauthor{\bsnm{Zhang},~\bfnm{Jiani}\binits{J.}},
  \bauthor{\bsnm{Srinivasan},~\bfnm{Balasubramaniam}\binits{B.}},
  \bauthor{\bsnm{Shen},~\bfnm{Zhengyuan}\binits{Z.}},
  \bauthor{\bsnm{Qin},~\bfnm{Xiao}\binits{X.}},
  \bauthor{\bsnm{Faloutsos},~\bfnm{Christos}\binits{C.}},
  \bauthor{\bsnm{Rangwala},~\bfnm{Huzefa}\binits{H.}} \AND
  \bauthor{\bsnm{Karypis},~\bfnm{George}\binits{G.}}
(\byear{2023}b).
\btitle{Mixed-Type Tabular Data Synthesis with Score-based Diffusion in Latent
  Space}.
\bjournal{arXiv preprint arXiv:2310.09656}.
\end{barticle}
\endbibitem

\bibitem[\protect\citeauthoryear{Zheng and
  Charoenphakdee}{2022}]{zheng2022diffusion}
\begin{barticle}[author]
\bauthor{\bsnm{Zheng},~\bfnm{Shuhan}\binits{S.}} \AND
  \bauthor{\bsnm{Charoenphakdee},~\bfnm{Nontawat}\binits{N.}}
(\byear{2022}).
\btitle{Diffusion models for missing value imputation in tabular data}.
\bjournal{arXiv preprint arXiv:2210.17128}.
\end{barticle}
\endbibitem

\end{thebibliography}





\end{document}